\documentclass[11pt, letterpaper]{arxiv}
\usepackage[all]{hypcap}
\usepackage[comma,numbers,sort,compress]{natbib}
\usepackage{hyperref}[citecolor=magenta]
\usepackage[capitalize,noabbrev]{cleveref}
\hypersetup{
    colorlinks = true,
    citecolor = {magenta},
}

\usepackage{microtype}
\usepackage{graphicx}
\usepackage{subfigure}
\usepackage{booktabs} 

\usepackage{xcolor}
\usepackage{colortbl}
\usepackage{algorithm}
\usepackage{algpseudocode}
\usepackage{setspace}

\usepackage{amsmath}
\usepackage{amssymb}
\usepackage{mathtools}
\usepackage{amsthm}

\usepackage{placeins}

\usepackage{fleqn, tabularx}

\usepackage{float}

\usepackage[most,skins,theorems]{tcolorbox}
\tcbset{
  aibox/.style={
    width=\linewidth,
    top=7pt,
    bottom=2pt,
    colback=rliableolive!13!white,
    colframe=black,
    colbacktitle=black,
    enhanced,
    center,
    attach boxed title to top left={yshift=-0.1in,xshift=0.15in},
    boxed title style={boxrule=0pt,colframe=white,},
  }
}
\newtcolorbox{AIbox}[2][]{aibox,title=#2,#1}

\usepackage{wrapfig}
\definecolor{lightblue}{rgb}{0.22,0.45,0.70}
\usepackage{fleqn, tabularx}

\definecolor{rliableolive}{HTML}{BBCC33}
\definecolor{rliableblue}{HTML}{77AADD}
\definecolor{rliablered}{HTML}{EE8866}

\usepackage{crossreftools}
\usepackage[suppress]{color-edits}

\usepackage{dsfont}
\usepackage{nicefrac}
\newtcolorbox{analysisbox}[1][]{
    enhanced jigsaw,
    colback=white,
    colframe=blue!75!black,
    fonttitle=\bfseries,
    boxsep=5pt,
    left=5pt,
    right=5pt,
    top=5pt,
    bottom=5pt,
    title=#1,
}
\definecolor{editInitialResponse}{RGB}{255, 235, 156} 
\definecolor{editBacktrack}{RGB}{0, 0, 139}
\definecolor{editRevisedResponse}{RGB}{255, 182, 193}

\usepackage{pifont}
\usepackage{soul}
\definecolor{highlightmistake}{RGB}{255, 179, 179} 
\definecolor{highlightcorrect}{RGB}{179, 255, 179}

\usepackage[capitalize,noabbrev]{cleveref}

\theoremstyle{plain}

\theoremstyle{definition}

\theoremstyle{remark}

\usepackage[textsize=tiny]{todonotes}
\usepackage{multirow}
\usepackage{enumitem}

\usepackage{amsmath,amsfonts,bm}

\def\eqref#1{Eq.~\ref{#1}}

\def\1{\bm{1}}

\DeclareMathAlphabet{\mathsfit}{\encodingdefault}{\sfdefault}{m}{sl}
\SetMathAlphabet{\mathsfit}{bold}{\encodingdefault}{\sfdefault}{bx}{n}

\newcommand{\bz}{\mathbf{z}}

\newcommand{\by}{\mathbf{y}}
\newcommand{\tby}{\widetilde{\mathbf{y}}}
\newcommand{\bx}{\mathbf{x}}

\newcommand{\solgen}{\pi^\mathrm{sol}_\theta}
\newcommand{\absgen}{\pi^\mathrm{abs}_\theta}

\newtcolorbox{promptbox}[2][]{  
listing only,
enhanced,
breakable,
colback=rliableolive!13!white,
colframe=black,
fontupper=\ttfamily,
title=#2,
#1}

\newcommand{\methodname}{{\texttt{RLAD}}}

\title{\methodname: Training LLMs to Discover Abstractions for Solving Reasoning Problems}

\author[*1]{Yuxiao Qu}
\author[*2]{Anikait Singh}
\author[*2]{Yoonho Lee}
\author[1]{Amrith Setlur}
\author[1]{Ruslan Salakhutdinov}
\author[2]{Chelsea Finn}
\author[1]{Aviral Kumar}

\affil[1]{Carnegie Mellon University}
\affil[2]{Stanford University}
\affil[*]{Equal contribution}

\correspondingauthor{yuxiaoq@andrew.cmu.edu, anikait@stanford.edu, yoonho@cs.stanford.edu. \newline \textbf{Project Website:} \textnormal{\url{https://cohenqu.github.io/rlad.github.io/}}}

\begin{document}

\maketitle

\begin{figure}[ht]
\centering
    \vspace{-0.5cm}
\includegraphics[width=\linewidth]{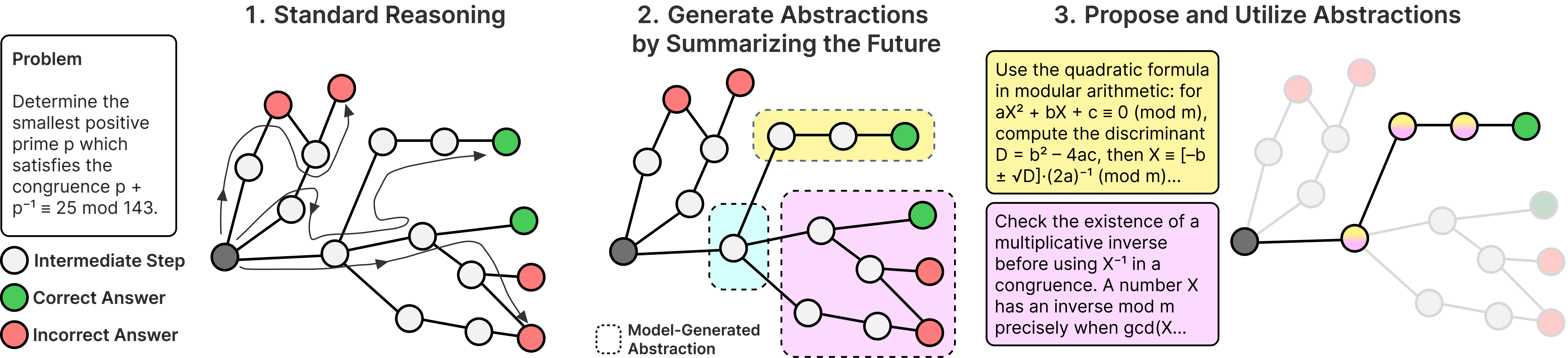}
\vspace{-5mm}
\caption{\label{fig:hint_category}
\footnotesize{\textbf{\emph{Reasoning abstractions illustrated in the solution-space graph for a problem.}} We depict the solution space as a graph of intermediate steps leading to correct or incorrect answers. \textbf{(1)} Standard reasoning explores this space along one sequential chain. \textbf{(2)} We generate textual abstractions by summarizing which intermediate steps led to which outcomes. \textbf{(3) }Such abstractions can be reused to guide reasoning more efficiently.}}
\end{figure}

{\textbf{Abstract:}} Reasoning requires going beyond pattern matching or memorization of solutions to identify and implement ``algorithmic procedures'' that can be used to deduce answers to hard problems. Doing so requires realizing the most relevant primitives, intermediate results, or shared procedures, and building upon them. While RL post-training on long chains of thought ultimately aims to uncover this kind of algorithmic behavior, most reasoning traces learned by large models fail to consistently capture or reuse procedures, instead drifting into verbose and degenerate exploration. To address more effective reasoning, we introduce \emph{reasoning abstractions}: concise natural language descriptions of procedural and factual knowledge that guide the model toward learning successful reasoning. We train models to be capable of proposing multiple abstractions given a problem, followed by RL that incentivizes building a solution while using the information provided by these abstractions. This results in a two-player RL training paradigm, abbreviated as \methodname, that jointly trains an abstraction generator and a solution generator. This setup effectively enables structured exploration, decouples learning signals of abstraction proposal and solution generation, and improves generalization to harder problems. We also show that allocating more test-time compute to generating abstractions is more beneficial for performance than generating more solutions at large test budgets, illustrating the role of abstractions in guiding meaningful exploration.


\vspace{-0.1cm}
\section{Introduction}
\label{intro}
\vspace{-0.2cm}

Modern machinery for training large language models (LLMs) to reason relies on incentivizing longer chains of thought via reinforcement learning (RL). This training approach largely incentivizes ``depth'': subsequent training iterations increase response length by incorporating new operations that usually attempt to verify or build on top of the line of reasoning being already pursued by the model~\citep{e3_extrapolation_2025}. This often results in very long chains of thought that appear to explore the solution search space, but degenerate into frequent logic switches and degenerate exploration (also referred to as ``underthinking''~\citep{wang2025thoughts}). One way to avoid this issue altogether is to directly optimize for ``breadth'': train language models to explore a diverse array of solution strategies, rather than committing to a seemingly good strategy right away and does not budge from it as more test-time compute is spent~\citep{yu2025dapo,yue2025does}. 

\textbf{\emph{How can we make models explore a breadth of reasoning strategies for a given problem?}} Abstractly, the most natural approach to do so is to train models to hypothesize new strategies to attack difficult problems and then attempt to utilize these strategies in the solution. We can do this by making models capable of discovering \emph{\textbf{reasoning abstractions}}: compressed representations of shared procedures that underlie multiple candidate solutions to a problem. For example, in math reasoning, such abstractions might correspond to useful intermediate lemmas or even some intermediate steps that do not succeed but illustrate what not to do. When presented in context, these abstractions function akin to ``hints'' on an exam, enabling LLMs to solve harder problems by building on the insights appearing in the abstraction. That is, when conditioned on abstractions, training via RL should train the LLM to implement useful meta strategies that utilize and compose the procedural information in the abstraction as best as possible to solve the problem, rather than attempting to search over the procedural information itself. This naturally boosts the diversity of solution strategies and behaviors that a model learns to utilize when encountering an unseen problem, in contrast to committing to a narrow set of approaches. In RL terminology, abstractions serve as high-level subgoals, skills, or priors---any of them depending upon context---guiding the low-level solution-generating policy.

In this work, we imbue LLMs with the capability of proposing and utilizing abstractions for solving problems. Concretely, we build reasoning models that, first, given an input problem, propose one or more reasoning abstractions. Subsequently, they generate a solution that utilizes the information and principles prescribed by these abstractions. To achieve this, we jointly train two LLMs via RL post-training: \textbf{(1)} an abstraction generator, and \textbf{(2)} an abstraction-conditioned solution generator. The abstraction generator is rewarded for the improvement in the accuracy of the solution generator, stemming from conditioning on the abstractions it proposes. The solution generator is rewarded to maximize accuracy in solving a problem when using the abstraction. To obtain a good initialization for RL training, we warmstart the abstraction generator by running supervised fine-tuning (SFT) on paired problem-abstraction data obtained with the help of stronger models. Specifically, we generate abstractions by summarizing multiple candidate solutions to a problem and prompt an LLM to generate a couple diverse abstractions. Once trained, the abstraction generator does not utilize any guidance from a larger model.

Our main contribution is the notion of \textit{reasoning abstractions}, how they can be obtained, training procedures to amplify them via RL training, and an illustration of why and how they can be used to improve reasoning performance and exploration of the search space. Concretely, we build an approach to imbue LLMs with the capability of proposing abstractions, and evaluate the model on a variety of math-reasoning benchmarks, AIME 2025~\citep{AIME2025}, DeepScaleR Hard~\citep{e3_extrapolation_2025}, and AMC 2023. We find an average 44\% improvement over state-of-the-art long chain-of-thought RL approaches (i.e., DAPO~\citep{yu2025dapo}) on AIME 2025, and show a benefit from generating diverse abstractions over solution sampling.
\vspace{-0.2cm}
\section{Related Work}
\vspace{-0.2cm}


\textbf{Scaling test-time compute and exploration.} Recent work highlights the promise of scaling test-time compute in different ways. One approach involves parallel sampling: sampling multiple reasoning rollouts and then selecting a winner via a scoring rule~\citep{uesato2022solvingmathwordproblems,wang2023selfconsistencyimproveschainthought,10.3115/1219840.1219862,feng2024alphazeroliketreesearchguidelarge,2024arXiv240803314S, yao2023treethoughtsdeliberateproblem, hao2023reasoninglanguagemodelplanning,snell2024scaling}. A complementary line of work iteratively edits a single trace, attempting to implement some sort of a sequential search within a single solution trace~\citep{2023arXiv230317651M,2024arXiv240718219Q,qu2024recursive,kumar2024traininglanguagemodelsselfcorrect}. As such, the sequential approach performs a bit worse on harder problems~\citep{snell2024scaling,qu2025optimizing}, where it often gets trapped in strategies that seem optimal but aren't actually~\citep{pan2025learning}. Yet it still performs better than parallel search on easier and medium difficulty problems~\citep{snell2024scaling}. Our approach of proposing and leveraging abstractions enables a kind of a hybrid between sequential sampling and parallel sampling, guided by the proposed abstractions. Some concurrent work~\citep{pan2025learning} studies directly interleaving parallel and sequential samples, and while it is similar in motivation to us, it only distills this interleaved structure into the model and does not run RL training to optimize the parallel and sequential sampling procedures employed here. Prior work has also utilized hand-designed scaffolds to integrate multi-step evaluations of intermediate hypotheses into reasoning~\citep{yao2023reactsynergizingreasoningacting, ho2023largelanguagemodelsreasoning, hao2023reasoninglanguagemodelplanning, li2023camelcommunicativeagentsmind}. In contrast, we do not rely on pre-defined interfaces but learn to \emph{automatically} propose useful abstractions.

\textbf{Using prior knowledge for LLM reasoning.}
Several threads of work converge on the idea that \textit{textual artifacts} such as examples, plans, or prompts, can serve as reusable knowledge that steers LLM behavior. Existing retrieval-augmented generation (RAG) pipelines assume a static corpus, typically of human-written text, and focus on improving retrieval heuristics~\citep{lewis2020retrieval,borgeaud2022improving,trivedi2022interleaving,2024arXiv241020753V,anonymous2025optimizing,li2025searcho1agenticsearchenhancedlarge}. Many works use LLMs to learn or refine prompts, either in an input-agnostic fashion~\citep{zhou2022large,yang2023large,pryzant2023automatic,fernando2023promptbreeder} or through input-specific edits based on feedback~\citep{shinn2023reflexion,madaan2023self,gou2023critic,yuksekgonul2025optimizing,lin2025sleep}. Other related work explores the use of synthetic demonstrations~\citep{zelikman2022starbootstrappingreasoningreasoning}, scratchpads~\citep{nye2021workscratchpadsintermediatecomputation}, and memory-augmented agents~\citep{Sch_fer_2020} to encode prior problem-solving knowledge. Two recent works demonstrate that LLMs can accumulate and reuse their own experience across tasks~\citep{zhao2024expel,suzgun2025dynamic}.
While one can view our abstractions as a form of prior procedural and factual knowledge produced before the model's  attempt, this knowledge is \textbf{(a)} input-dependent and \textbf{(c)} is not acquired from an external source at deployment, but rather is ``proposed'' by the model itself. Imbuing models with this capability requires a cooperative RL training process. To our knowledge, prior work on textual artifacts do not quite train models to be capable of generating these artifacts on their own.
\vspace{-0.4cm}
\section{Preliminaries and Notation}
\vspace{-0.25cm}

We assume an LLM is provided access to a problem $\bx$, and generates a stream of tokens $\by$ that ends in an estimate of the answer. We assume access to a rule-based ground-truth 0/1 reward $\mathrm{Acc}_\bx(\by, \by^\star)$ $\in$ $\{0, 1\}$ that measures correctness of the produced answer $\by$, against the ground-truth solution $\by^\star$ for a question $\bx$. For training, we are given a dataset $\mathcal{D}_\textrm{train} = \{(\bx_i, \by_i^\star)\}_{i=1}^N$ of problems $\bx_i$ and solutions $\by_i^\star$ that end with the correct answer. Our goal is to train the LLM $\pi(\cdot | \bx)$ such that it
achieves high rewards on a test distribution of problems $ \mathcal{P}_\mathrm{test}$. We evaluate models by measuring the average accuracy under $\mathcal{P}_\mathrm{test}$. We also measure the pass@k metric, where for problem $\bx$, we sample $k$ solutions $\by_1, \ldots, \by_k \sim \pi(\cdot | \bx)$, and consider the problem to be solved if any of these $k$ traces is correct. This metric couples accuracy with diversity, i.e., it attains the largest value when the model effectively finds diverse, good responses. To reduce variance in estimating pass@k, we sample $n \geq k$ samples per problem and use the unbiased estimator introduced in Codex~\citep{chen2021evaluatinglargelanguagemodels}: $1-\binom{n-c}{k} / \binom{n}{k}$, where $c \leq n$ is the number of correct samples.
\vspace{-0.4cm}
\section{Reasoning Abstractions and Why They Are Useful}
\label{sec:reasoning_abstractions}
\vspace{-0.25cm}
Solving reasoning problems often requires composing both \emph{procedural} knowledge (e.g., how to apply a root-finding algorithm) and \emph{factual} knowledge (e.g., relevant lemmas or intermediate results). Current approaches train models to reason via reinforcement learning (RL) on long chains of thought. However, this is often ineffective as RL often tends to optimize for ``depth'', producing longer traces where each subsequent segment builds on the last segment (e.g., verifying prior calculations), rather than ``breadth'', i.e., exploring diverse solution strategies or utilizing seemingly irrelevant procedures when needed. We now introduce the concept of \textbf{\emph{reasoning abstractions}}, concise insights that explicitly encode a range of useful procedural and factual concepts for a problem. 

\vspace{-0.2cm}
\subsection{Proposing Good Reasoning Abstractions by Summarizing Solution Attempts}
\label{subsec:reasoning_abstractions}
\vspace{-0.2cm}
To imbue LLMs with the capability of proposing abstractions, we warmstart the LLM using a dataset consisting of problems paired with a small set of high-quality reasoning abstractions, created synthetically. Perhaps the most natural way to obtain an initial set of reasoning abstractions is to collect a diverse set of traces attempting to solve a problem and then summarize useful concepts appearing in these traces (see Figure~\ref{fig:hint_category} for an illustration). More formally, consider the space of possible reasoning traces for a given problem as a graph where the nodes of the graph represent intermediate states encountered when solving a question (see~\cref{fig:hint_category}). Good abstractions would identify useful substructures within this larger reasoning graph. For example, an abstraction can capture whether a set of strategies lead to a similar outcome or another set of tactics leads to an error being consistently made.

\textcolor{lightblue}{\textbf{Generating abstractions.}} We now quantitatively evaluate whether we can generate useful abstractions by summarizing key insights in reasoning and solution traces that solve a problem. To do so, we prompt a model to generate solution traces and prompt a stronger model to deduce useful patterns from the responses of the first model. Concretely, we utilize the Qwen3~\citep{qwen3_2025} series of models to produce solutions and a stronger reasoning model, o4-mini, to generate abstractions. While this approach is not perfect, it enables us to validate the feasibility of the concept of reasoning abstractions and generate warmstart data for \emph{training} LLMs to propose abstractions. To ensure that the abstractions do not ``leak'' content of the solution, we verify post-hoc that prompting a model with only the abstraction and no problem yields zero accuracy when sampling 16 times from the base model. This makes these abstractions suitable for our study as they only provide useful information while not allowing the model to shortcut to the answer. 

\textcolor{lightblue}{\textbf{Evaluating abstractions.}} To validate the efficacy of using abstractions, we adopt a simple test based on performance after conditional generation. Concretely, let us denote the LLM policy that produces a solution conditioned on the problem $\bx$ as $\pi_\theta^\mathrm{sol}(\cdot|\bx)$. A good abstraction $\bz$ is a sequence of tokens that provides some useful procedural and factual information to improve model performance:
\begin{align}
\label{eq:abstraction_success}
\mathbb{E}_{\tby \sim \solgen(\cdot|\bx, \bz)} \left[ \mathrm{Acc}(\tby, \by^*) \right] >
\mathbb{E}_{\tby \sim \solgen(\cdot|\bx)} \left[ \mathrm{Acc}(\tby, \by^*) \right].
\end{align}

\vspace{-0.5cm}
\subsection{Results and Observations}
\label{subsec:results_for_abstractions}
\vspace{-0.2cm}

\begin{wraptable}{r}{0.4\linewidth}
\vspace{-0.4cm}
\centering
\footnotesize
\setlength{\tabcolsep}{3pt}
\renewcommand{\arraystretch}{0.9}
\begin{tabular}{c cc cc}
\toprule
\multirow{2}{*}{\textbf{k}} &
\multicolumn{2}{c}{\textbf{pass@k}} &
\multicolumn{2}{c}{\textbf{coverage (max@k)}} \\
\cmidrule(lr){2-3} \cmidrule(lr){4-5}
& w/o abs & w/ abs & w/o abs & w/ abs \\
\midrule
1  & 14.0\% & \textbf{18.0\%} & 19.5\% & \textbf{22.5\%} \\
2  & 17.5\% & \textbf{22.5\%} & 24.0\% & \textbf{28.5\%} \\
4  & 20.0\% & \textbf{26.5\%} & 28.5\% & \textbf{34.0\%} \\
8  & 22.8\% & \textbf{30.2\%} & 31.8\% & \textbf{38.8\%} \\
16 & 24.7\% & \textbf{33.2\%} & 35.0\% & \textbf{42.9\%} \\
\bottomrule
\end{tabular}
\vspace{-0.2cm}
\caption{\footnotesize{\emph{\textbf{pass@k accuracy and max@k coverage on the ARC-AGI benchmark.}} Abstractions yield consistent gains in both metrics.}}
\label{tab:arc_passk}
\vspace{-0.3cm}
\end{wraptable}

\begin{figure}
\vspace{-0.3cm}
\centering
\includegraphics[width=0.95\linewidth]{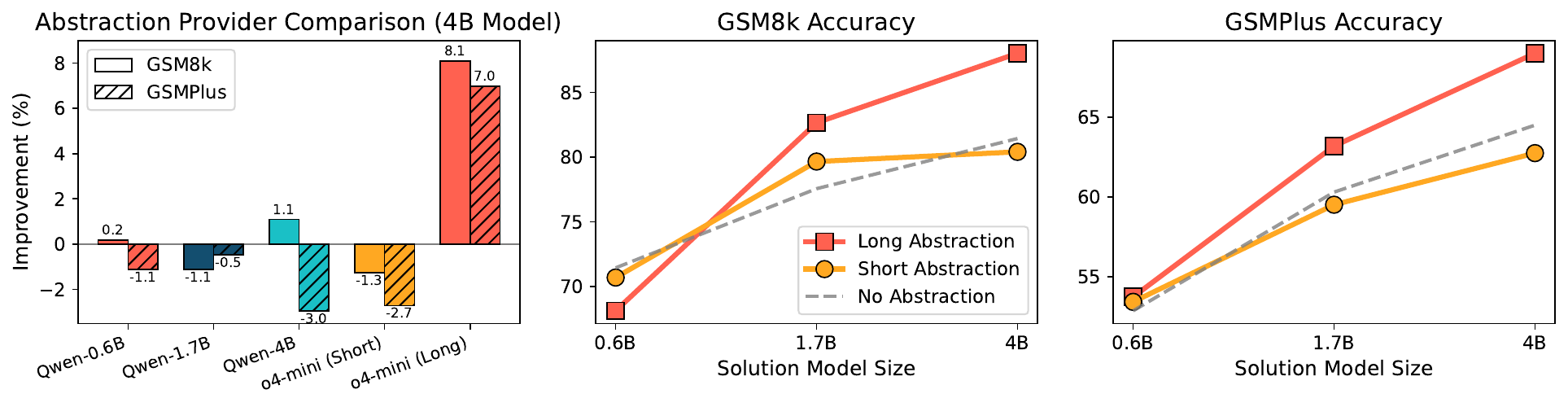}
\vspace{-0.5cm}
\caption{\label{fig:grid_plot}
\footnotesize{\textit{\textbf{Benefits from abstractions rely crucially on strength of the solver, abstraction length, and solution model}}. Most configurations fail to yield gains; only {o4-mini} with long and detailed abstractions shows consistent improvements across the GSM8k and GSMPlus datasets (left). The capability of the problem solver conditioned on abstractions also matters: even strong abstractions help only if the solution model is sufficiently capable (middle, right).
}}
\vspace{-0.4cm}
\end{figure}

\textcolor{lightblue}{\textbf{Evaluation on math reasoning.}} After generating abstractions, we measure their quality by evaluating Equation~\ref{eq:abstraction_success}, i.e., by checking if conditioning the solution generator on a set of abstractions improves its accuracy. Results in~\cref{fig:grid_plot} show that conditioning a problem solver on abstractions improves accuracy when two conditions hold simultaneously: (i) the abstraction is not too short (e.g., not just a few words that are not informative; example in Appendix~\ref{sec:short}) and is generated by a strong generator ({o4-mini}) and (ii) the solution generator has sufficient instruction-following capability ({Qwen3-1.7B} or {Qwen3-4B}) of interpreting and utilizing the generated abstraction. These results confirm that good abstractions (satisfying~\eqref{eq:abstraction_success}) can be obtained for math problems, but neither the ability to generate them nor the ability to leverage them in solutions arises naturally. In~\cref{sec:method}, we will describe our method for explicitly training models to propose and use such abstractions effectively.

\textcolor{lightblue}{\textbf{Evaluation on ARC-AGI.}} We also evaluate abstractions on the ARC-AGI benchmark. We present some details of our setup in Appendix~\ref{sec:full_details_nonmath}. We evaluate on 90 ARC puzzles evenly derived from the test sets of ARC-AGI 1, ARC-AGI 2, and BARC~\citep{li2024combininginductiontransductionabstract}. In Table~\ref{tab:arc_passk}, we present the pass@k and coverage performance (\% of unit tests successfully passed) of the base Qwen3-4B  model when conditioned on a proposed abstraction vs not using any abstraction. We see a positive improvement in both metrics on this domain over multiple samples, indicating an improvement from utilizing reasoning abstractions.

\textcolor{lightblue}{\textbf{Interpreting the generated abstractions.}} We also show some examples of the discovered abstractions in Appendix~\ref{sec:abs_category}. We now attempt to interpret these abstractions by classifying them into several categories. We observe that the discovered abstractions often correspond to useful techniques (e.g., ``launchpoint'' in Appendix~\ref{sec:abs_category}), a useful lemma or heuristic principle (e.g., ``blind-follow'' in Appendix~\ref{sec:abs_category}), and  cautionary examples that demonstrate common pitfalls encountered when solving a problem (e.g., ``caution alert'' in Appendix~\ref{sec:abs_category}). These abstractions distill complex reasoning patterns and potential approaches into useful nuggets, allowing models to generalize across structurally similar problems.
Finally, we would like to remind the reader and emphasize that the above qualitative results from interpreting the discovered abstractions are specific to an individual problem, and not representative of the process being used to discover them. Our approach for generating abstractions is neither hand-engineered for interpretability nor does it use any such heuristics in addition to the summarization procedure discussed above.
 

\textcolor{lightblue}{\textbf{Good abstractions exist in many domains.}} 
We also find that the summarization approach can be used to identify a set of useful reasoning abstractions on many domains, including healthcare, human behavior, legal reasoning, and web security. Of course, the proportion of an abstraction devoted to procedural knowledge and factual knowledge is different in these domains compared to math. That said, we find that using reasoning abstractions improves performance by 30\% on average over 37 tasks from RAFT~\citep{alex2021raft}, CLUES~\citep{menon2022clues}, and LegalBench~\citep{guha2023legalbench}. We show some examples in~\cref{fig:abstraction_examples} and full results in Appendix~\ref{sec:full_details_nonmath}.

\begin{AIbox}{Takeaway: Reasoning abstractions summarize insights useful for guiding solution traces}
Reasoning abstractions summarize procedural and factual knowledge that is useful for learning to solve problems via diverse strategies. Proposing abstractions generated by summarizing solution traces already improves performance of base generators by 30\% for math reasoning.
\end{AIbox}

\vspace{-0.2cm}
\section{\methodname: Learning to Propose Reasoning Abstractions}
\vspace{-0.2cm}
\label{sec:method}

Having defined the notion of reasoning abstractions and shown that they can improve performance when adhered to for reasoning, we will now develop an approach to train LLMs to be capable of both proposing and utilizing abstractions. Doing so requires training an \emph{abstraction generator}: an LLM, $\bz \sim \absgen(\cdot|\bx)$ that proposes candidate abstractions $\bz$ given problem $\bx$, and an abstraction-conditioned solution generator, $\by \sim \solgen(\cdot|\bx, \bz)$, that produces a solution $\by$ given $\bx$ and abstraction $\bz$. Note that $\bz$ is parameterized as a variable-length sequence of tokens and might consist of one or more facts or procedures. While our approach applies to the case when $\absgen$ produces more than one abstraction, we abuse notation and subsume multiple abstractions into one to avoid clutter. We now describe RL with abstraction discovery~(\methodname), our method for training these models via RL.

\vspace{-0.2cm}
\subsection{Training $\absgen$ and $\solgen$ via RL}
\label{subsec:method_main}
\vspace{-0.1cm}

\begin{figure}
\centering
\vspace{-0.3cm}
\includegraphics[width=0.95\linewidth]{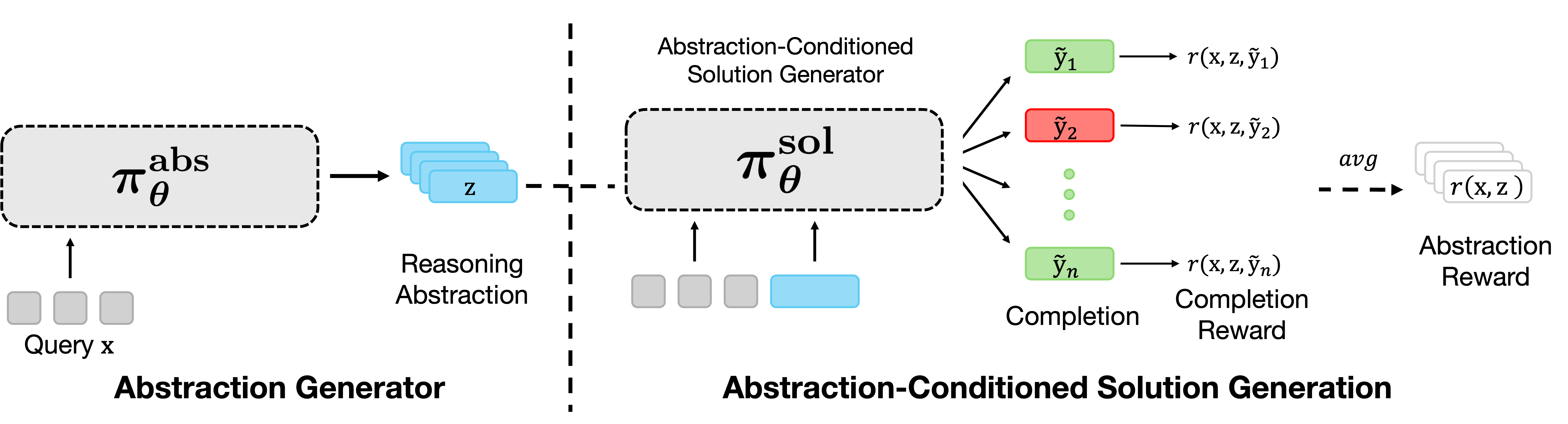}
\vspace{-0.2cm}
\caption{\footnotesize{\textit{\textbf{\methodname{} training paradigm.}}  We train an abstraction generator, $\absgen$, that proposes some reasoning abstractions  conditioned on the question $\bx$, denoted as $\bz$. Then, the solution generator, $\solgen$, is trained to produce a response, $\tilde{\by}$, conditioned on the generated abstraction $\bz$. The reward used for training $\absgen$ corresponds to the average success rate of the solution generator conditioned on the proposed abstraction.}}
\label{fig:workflow}
\vspace{-0.4cm}
\end{figure}
The core principle behind our approach is that an abstraction $\bz$ is successful at a given problem $\bx$ if it can maximally help $\solgen(\cdot|\bx, \bz)$ find correct responses to the question $\bx$, without actually leaking the answer itself. To ensure this, we explicitly prompt the model not to reveal solution information, and we further use an LLM-based judge to verify that the generated guidance does not contain the answer.
To convert this into an RL objective, we design a reward function that rewards an abstraction $\bz$ with the expected success of solutions generated by $\solgen$ conditioned on $\bz$:
\begin{align}
    \label{eq:reward_absgen_naive}
    r_{\solgen}(\bx, \bz) := \mathbb{E}_{\tby \sim \solgen(\cdot|\bx, \bz)} \left[ \mathrm{Acc}_\bx(\tby, \by^*) \right],
\end{align}
where $\by^*$ is the ground-truth answer and $\mathrm{Acc}_\bx(\cdot, \cdot)$ denotes the 0/1 accuracy on problem $\bx$. To train $\solgen$, one can then adopt the fairly straightforward approach of maximizing 0/1 binary outcome reward, now conditioned on a given abstraction $\bz$ sampled previously from $\absgen$, akin to recent results RL~\citep{deepseekai2025deepseekr1incentivizingreasoningcapability}. 
Formally, we set the reward for a solution as:  $r(\bx, \bz, \tby) := \mathrm{Acc}_\bx(\tby, \by^*)$. With these reward functions in place, perhaps the most natural approach then would be to train $\absgen$ to maximize $r_{\solgen}$ for a fixed $\solgen$ on a dataset of prompts $\mathcal{D}_{\absgen}$, while also iteratively training $\solgen$ to maximize the reward function $r$ on modified prompts generated by concatenating a set of sampled abstraction $\bz$ on a dataset of problems, $\mathcal{D}_{\solgen}$. This maximization could be done via on-policy RL (e.g., GRPO~\citep{shao2024deepseekmath}) or offline RL methods (e.g., DPO~\citep{rafailov2023direct}, STaR~\citep{zelikman2022star}). This represents a cooperative two-player game.

\textbf{Challenges with na\"ive reward design.} While the approach so far is extremely simple, it presents some challenges. In particular, the reward functions defined above can result in undesirable solutions in a rather nuanced manner: \textbf{(1)} if $\absgen$ learns to solve problem $\bx$ in itself, it will still be rewarded highly by $r_{\solgen}$ but is not a desirable abstraction; \textbf{(2)} if $\solgen$ is too weak or too strong, such that it is either able to always solve the problem $\bx$ or never solves it, then $r_{\solgen}$ will not provide a meaningful signal to update $\absgen$; and \textbf{(3)} similar to the above failure modes, training $\solgen$ via on-policy RL may result in it ignoring the abstraction $\bz$ altogether no matter how useful it is. Abstractly, all of these challenges stem from an asymmetry in the strength of $\absgen$ and $\solgen$, where one may drown out the learning signal for the other. We therefore build a modified reward system for training.

\textbf{Modifying reward design.} To address these potential failure modes, we make a small but consequential change to the training reward system inspired by \citet{kumar2024traininglanguagemodelsselfcorrect}. In particular, we train $\solgen$ on a mixture of prompts $\bx$ augmented by abstractions $\bz$ and prompts $\bx$ without any abstractions at all. In this process, while we utilize $\mathrm{Acc}_\bx$ as discussed above on a given response, we simply zero out rewards for any trace generated on $\bx$ without abstractions. When utilizing KL-constrained RL fine-tuning, $\solgen$ is now trained to closely mimic the distribution of responses as the reference LLM on questions $\bx$ but must attempt to find ways to optimize reward on the same question $\bx$ when augmented with an abstraction. This can be accomplished only when $\solgen$ learns to utilize the provided abstraction carefully, hence addressing one of the challenges above.
Formally, the updated versions of these reward functions are:
\begin{align}
    \label{eq:reward_func_real}
    r(\bx, \bz, \tby) &:=
    \begin{cases}
    0, & \text{if } \bz = \emptyset \\
    \mathrm{Acc}_{\bx}(\tby, \by^*), & \text{otherwise} \\
    \end{cases} \\
    r_{\solgen}(\bx, \bz) &:= \mathbb{E}_{\tby \sim \solgen(\cdot|\bx, \bz)} [\mathrm{Acc}_\bx(\tby, \by^*)].
\vspace{-0.2cm}
\end{align}

\vspace{-0.2cm}
\subsection{Warmstarting $\absgen$ from a Good Initialization}
\vspace{-0.1cm}
While the above approach prescribes a recipe for RL training of $\absgen$ and $\solgen$, any such recipe critically relies on the ability of the initialized model to generate somewhat meaningful abstractions and meaningful solutions conditioned on the abstraction input, respectively, right from the beginning of RL training. Inspired by the approach of running an initial phase of SFT to imbue into the model the basic structure of a long chain-of-thought before RL~\citep{deepseekai2025deepseekr1incentivizingreasoningcapability,qu2025optimizing}, we run an initial phase of SFT to imbue into $\absgen$ the basic capabilities of producing abstractions. We did not perform warmstart training for $\solgen$ as we find that the base Qwen3 models we were utilizing were capable of utilizing abstractions well. For this initial warmstart phase, we follow the protocol from Section~\ref{sec:reasoning_abstractions} and construct a corpus $\{(\bx_i, \bz_i, \by_i)\}_{i=1}^M$ by prompting strong models. For each training problem-solution pair $(\bx, \by^*)$ in our training set, we first generate an abstraction $\bz$ via summarization, discarding any that leak $\by^*$. 

\begin{figure} \centering
\vspace{-3mm}
\includegraphics[width=\linewidth]{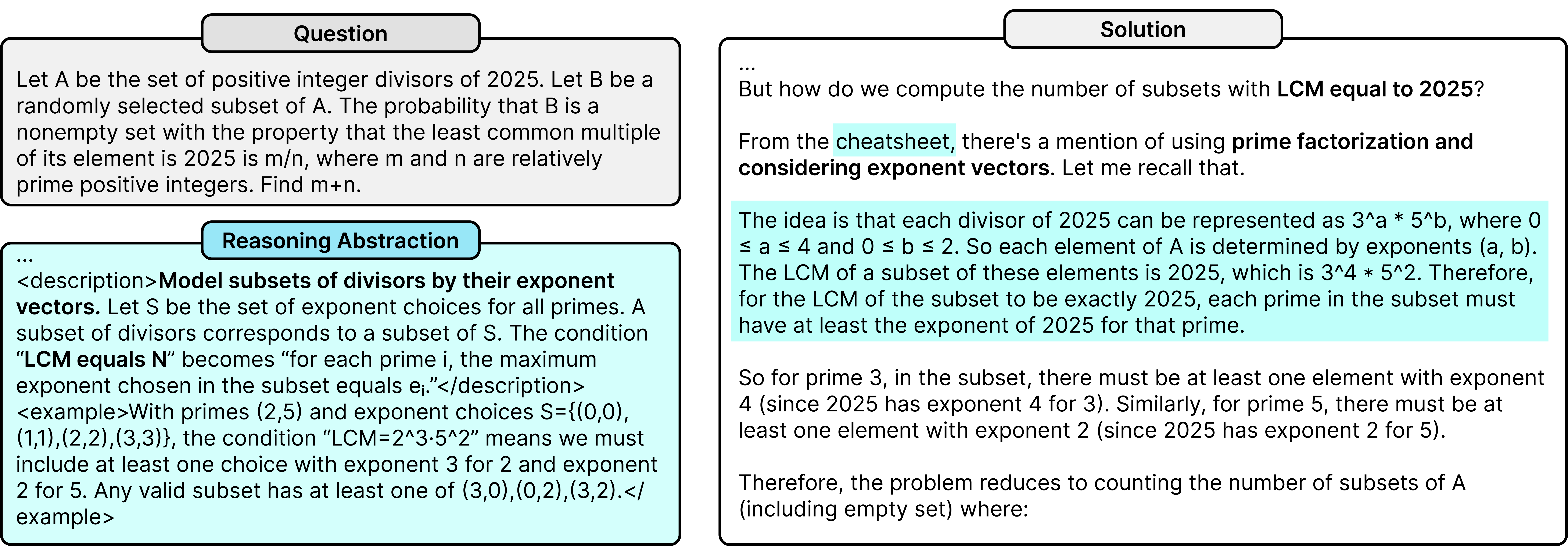}
\vspace{-6mm}
\caption{\footnotesize \textit{\textbf{A typical example of a reasoning abstraction proposed by our abstraction generator.}} In the solution, we see (\textbf{in blue}) references to the abstraction (``cheatsheet'') and keywords from the abstraction being used meaningfully in the reasoning trace of the solution generator model.}
\vspace{-4mm}
\label{fig:math-abstraction-example}
\end{figure}

\textbf{Practical approach and algorithm details.} For warmstarting the abstraction generator, we utilize abstractions generated by {o4-mini}. We then use a weaker model (GPT 4.1-mini) to check the efficacy of each abstraction when conditioned on by comparing the success rate of the solution generator with and without an abstraction. We filter abstractions that don't result in an increase in solution generation performance to form our seed set of abstractions. Then, we run supervised fine-tuning (SFT) for $5$ epochs on the seed dataset to obtain an initial abstraction generator. For solution generation, we directly start from Qwen3-1.7B~\citep{qwen3_2025}, a 1.7B reasoning model distilled from Qwen3-32B. 

After SFT, we employ \methodname{} to fine-tune the abstraction generator and abstraction-conditioned solution generator via RL. For the abstraction generator, we opt to use ``batched'' offline RL via RFT~\citep{yuan2023scaling}, since reward computation by rolling out the solution generator the on the fly and running online RL was infeasible using our RL infrastructure and within the compute we had access to. To train the solution generator, we build on the RL recipe from DAPO~\citep{yu2025dapo}, and include token-level policy loss normalization and asymmetric clipping. Building upon implementation of \citet{e3_extrapolation_2025}, we employ a two stage curriculum where we partition the DeepScaleR~\citep{deepscaler2025} mixture by success rate of the base model into three sets: (1) easy, (2) medium, and (3) hard, 
where we fine-tune first on easy problems with an 8K token budget and then on medium problems. We utilize the hard split as a held-out, evaluation subset, which we denote as {``\textbf{DeepScaleR [Hard]}''}. Note that these splits are different from the \texttt{e3}~\citep{e3_extrapolation_2025} work, while the principle is inspired by that work. We outline hyperparameters and details in Appendix~\ref{app:exp_details} and provide a pseudocode in Algorithm~\ref{alg:abstraction}.

\begin{AIbox}{Summary: \methodname{} method design}
\methodname{} jointly optimizes the abstraction generator $\absgen$ and solution generator $\solgen$ with RL, using reward functions in Equation~\ref{eq:reward_func_real}. These reward functions incentivize $\solgen$ to utilize abstractions and incentivize $\absgen$ to propose useful abstractions per problem.
\end{AIbox}

\vspace{-0.3cm}
\section{Experimental Evaluation}
\vspace{-0.25cm}
The goal of our experiments is to evaluate the efficacy of \methodname{} in improving the reasoning capabilities of LLMs through abstraction-guided solution generation. Specifically, we aim to answer the following questions: \textbf{(1)} Does \methodname{} improve pass@1 accuracy across several reasoning benchmarks compared to direct solution generation?, \textbf{(2)} How does \methodname{} scale as more abstractions and solutions are generated?, and \textbf{(3)} What makes the generated abstractions useful, how faithfully are they followed, and how do they guide and improve solution generation? We compare \methodname{} with strong models on three math reasoning benchmarks: AMC 2023, AIME 2025, and DeepScaleR Hard~\citep{deepscaler2025}, which itself is a subset of hard problems from the OmniMATH mixture on which DeepSeek-R1 distilled Qwen-32B model attains an accuracy of $\leq 10\%$. We also fine-tune an abstraction generator for the ARC-AGI program synthesis tasks, and conduct a similar comparison on 90 ARC puzzles evenly derived from the test sets of ARC-AGI 1, ARC-AGI 2, and BARC~\citep{li2024combininginductiontransductionabstract}. Finally, we also perform several ablations to better understand abstractions produced by \methodname{} and discuss them later.

\vspace{-0.25cm}
\subsection{Main Performance Results on Math Reasoning Benchmarks}
\vspace{-0.2cm}
We evaluate \methodname{} under three settings: (1) \textbf{w/o abs}, without abstractions; (2) \textbf{w/ abs (avg)}, average performance over generations conditioned on 4 abstractions per problem; and (3) \textbf{w/ abs (best)}: using the best-performing abstraction (in a set of 4 proposed abstractions per problem).
We observe that \methodname{} outperforms the base model and variant fine-tuned with RL on the same prompts via DAPO~\citep{yu2025dapo} without abstractions, across all settings and benchmarks~(\cref{tab:main_results_correct}). This highlights that \methodname{} can propose and leverage abstractions to improve reasoning performance. Interestingly, these performance gains are not limited to abstraction-conditioned inference: even in the \textbf{w/o abs} setting, where no abstraction is provided during inference, \methodname{} improves over the prior methods, when trained with abstractions via \methodname{}. This suggests that exposure to diverse abstractions during training enhances the model's general capability on these training problems. We observe similar trends on additional benchmarks, including AIME 2024 and HMMT 2025 (see Appendix~\ref{sec:AIME24_HMMT25}), where \methodname{} improves in the w/o abs setting. 
Finally, in Appendix~\ref{app:abstractions_examples}, we also measure the performance of \methodname{} when different token budgets are allowed for reasoning -- while Table~\ref{tab:main_results_correct} measures performance at a budget of 32K tokens, we also measure performance at 8K and 16K budgets and find \methodname{} to be more effective  than other approaches.

\begin{table*}[t]
  \centering
  \footnotesize
  \newcolumntype{C}{>{\centering\arraybackslash}X}
  \setlength{\tabcolsep}{4pt}
  \begin{tabularx}{\linewidth}{l CCC CCC CCC}
    \toprule
    \multirow{2}{*}{\textbf{Approach}} &
    \multicolumn{3}{c}{\textbf{AIME 2025}} &
    \multicolumn{3}{c}{\textbf{DeepScaleR [Hard]}} &
    \multicolumn{3}{c}{\textbf{AMC 2023}} \\
    \cmidrule(lr){2-4} \cmidrule(lr){5-7} \cmidrule(lr){8-10}
    & w/o abs (avg) & w/ abs (avg) & w/ abs (best) & w/o abs (avg) & w/ abs (avg) & w/ abs (best) & w/o abs (avg) & w/ abs (avg) & w/ abs (best) \\
    \midrule
    Qwen-3-1.7B & 33.75 & 36.25 & 40.00 & 20.21 & 22.14 & 32.50 & 86.41 & 78.01 & 84.53 \\
    + DAPO      & 37.92 & 34.90 & 39.79 & 21.67 & 21.88 & 33.54 & 86.41 & 81.99 & 88.44 \\
    + \textbf{\methodname{} (Ours)} & 38.04 & \textbf{42.45} & \textbf{48.33} & 23.54 & \textbf{24.84} & \textbf{35.54} & 87.25 & \textbf{88.35} & \textbf{91.72} \\
    \bottomrule
  \end{tabularx}
  \vspace{-0.2cm}
  \caption{\footnotesize \textbf{Accuracy on math reasoning benchmarks.} We measure test performance on three benchmarks (AIME 2025, DeepScaleR Hard, and AMC 2023) with the base Qwen 3-1.7B model, DAPO, and \methodname{}. Performance is measured under three settings: \textbf{(1) w/o abs}, without abstractions; \textbf{(2) w/ abs (avg)}, average accuracy over solutions generated when conditioned on four proposed abstractions per problem; and  \textbf{(3) w/ abs (best)}, using the best-performing abstraction among the four generated in (2). \methodname{} achieves consistent gains across both abstraction-conditioned and no-abstraction settings.}
  \vspace{-0.4cm}
  \label{tab:main_results_correct}
\end{table*}

\vspace{-0.3cm}
\subsection{Understanding Properties of \methodname{}}
\vspace{-0.2cm}

In this section, we conduct a number of additional experiments to understand the behavior of \methodname{} in regards to the usefulness of the proposed abstractions and the performance of the \methodname{} algorithm when provided with a large budget on the total inference compute.

\textcolor{lightblue}{\textbf{1) ``Weak-to-strong'' generalization of the abstraction generator.}} 
We next evaluate the weak-to-strong generalization of our method by pairing our trained abstraction generator with {o4-mini} as the solution generator. We use a fixed 24k generation token budget and 4 samples per question. Without abstractions, {o4-mini} achieves 80.38\% pass@1, 82.26\% pass@2, and 84.77\% pass@4. Conditioning on both the problem and the proposed abstractions improves performance to 85.83\% pass@1, 88.33\% pass@2, and 90.00\% pass@4 accuracy. Thus, conditioning on abstractions consistently yields higher pass@k accuracy compared to question-only conditioning, even for this strong reasoning model. These gains demonstrate that abstractions, though produced by a comparatively weaker model, can transfer effectively to a stronger solution generator, providing evidence that abstractions generalize in a weak-to-strong setting and enables downstream improvements without additional supervision or modifications to the strong model.

\textcolor{lightblue}{\textbf{2) Compute tradeoffs b/w abstraction and solution generation.}}
Next, we study how to allocate compute between generating more abstractions and sampling solutions to attain maximal performance within a given inference budget. This corresponds to a ``compute-optimal strategy''~\citep{snell2024scaling} for partitioning compute between abstraction and solution generation. If the model typically fails by making small local errors in its computation, then additional concise abstractions may not help it as much as simply trying again to generate a locally similar solution (optimizing ``depth``). In contrast, if the model tends to pursue a seemingly plausible but incorrect approach and is unable to easily recover or switch approaches, then conditioning on diverse abstractions can help by offering alternative high-level approaches toward the correct answer. In other words, when the model has a tendency to explore ``depth'' over ``breadth'' of solution strategies, abstractions can help improve performance. With this intuition, we hypothesize that when the compute budget permits only a limited number of samples, allocating more compute to sampling multiple solutions but retaining only a few abstractions will improve success more. However, once pass@k for a single abstraction begins to saturate, performance gains are more likely to come from scaling the diversity of abstractions, which enables the model to explore different strategies.

\begin{figure}
    \centering
    \vspace{-0.3cm}
    \includegraphics[width=0.99\linewidth]{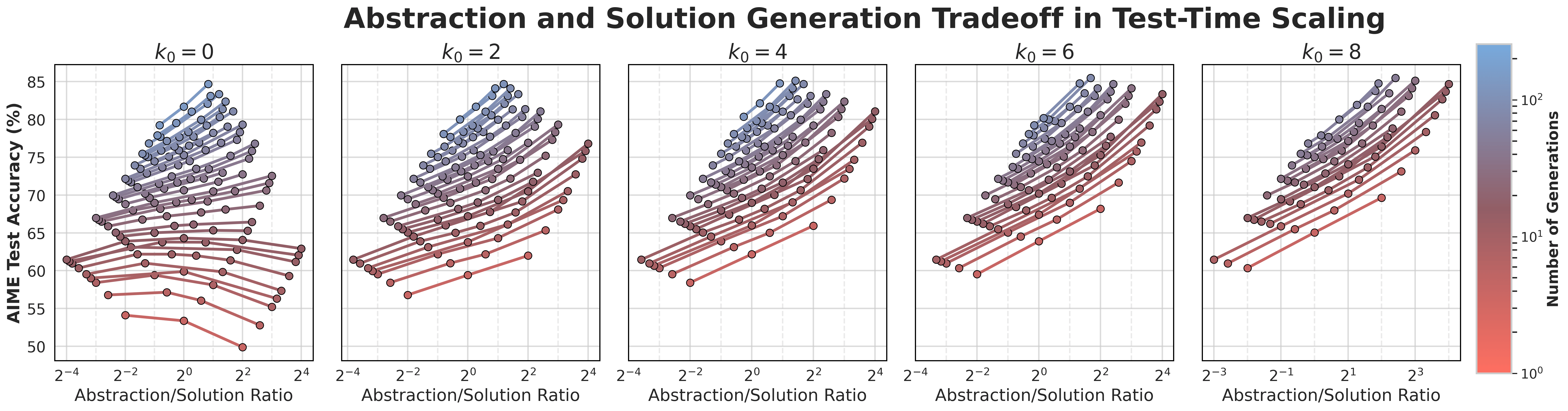}
    \vspace{-0.2cm}
    \caption{\footnotesize \emph{\textbf{Tradeoff of abstraction and solution generation on AIME 2025.}} As the total inference compute budget increases (color scheme on the right), we find better performance efficiency when allocating our budget to abstraction generation rather than solution generation, for all values of normalization offset $k_0$ given to us.}
    \label{fig:abstraction_ratio}
    \vspace{-4mm}
\end{figure}

To validate this hypothesis, we plot \emph{iso-compute} scaling curves under a fixed compute budget $\mathcal{C}$, we distribute between abstractions and solution generation. Specifically, we denote the number of abstractions as $m$ and the number of solutions sampled per abstraction as $k$, such that $m \times k = \mathcal{C}$.
To better isolate the utility of abstractions, we make several more projections of the iso-compute curve, each with a different ``normalization offset'' \( k_0 \). This normalization offset $k_0$ \emph{discounts} performance gains that stem from trying the problem again and making local modifications (e.g., small edits that do not require new strategy changes), but measures the performance gains that stem from major changes in content of a response. Concretely, for a non-zero $k_0$, we plot \emph{iso-compute} frontiers when $m \times (k-k_0) = \mathcal{C}$. This formulation captures the amount of compute that is spent on ``meaningful'' samples that go beyond the model's local neighborhood. Figure~\ref{fig:abstraction_ratio} shows these \emph{iso-compute} frontiers for different values of the compute budget $\mathcal{C}$. The x-axis plots the ratio between abstractions and adjusted solutions, \( m / (k - k_0) \). Each curve corresponds to a fixed total compute budget. 
In Figure~\ref{fig:abstraction_ratio},  across $k_0 \in \{0, 2, 4, 6, 8\}$, shifting compute toward abstractions consistently yields greater improvements than allocating the same additional compute to solution refinements, especially as the total compute budget increases. {\emph{\textcolor{lightblue}{\textbf{This supports the conclusion that}} once local errors in the chain-of-thought have been addressed, it is more effective to increase the diversity of strategies used via abstraction conditioning rather than to continuing to scale up CoTs alone.}

\begin{wraptable}{r}{0.7\linewidth}
\centering
\vspace{-0.3cm}
\begin{tabular}{lccccc}
\toprule
 & 1 & 4 & 16 & 64 & 256 \\
\midrule
DAPO, w/o abs (pass@$n^2$) & 0.37 & 0.51 & 0.65 & 0.77 & 0.82 \\
\methodname{}, w/ abs (pass@$n \times n$) & \textbf{0.41} & \textbf{0.59} & \textbf{0.71} & \textbf{0.80} & \textbf{0.87} \\
\bottomrule
\end{tabular}
\vspace{-0.1cm}
\caption{\footnotesize \textbf{Pass@k comparison under equal compute.} 
For each $n$, we compare $n^2$ solution samples (no abstractions) with 
$n$ abstractions $\times$ $n$ solutions per abstraction. Abstraction 
conditioning yields consistent improvements.}
\label{tab:passk_abs}
\vspace{-0.2cm}
\end{wraptable}
To obtain more evidence to support this result, we also evaluate pass@k of DAPO and \methodname{} for a fixed compute budget for both abstraction and non-abstraction generation on AIME 2025. Here, for a given value $n$, we compare two approaches: \textbf{(1)} sampling $n^2$ solutions without abstractions, and \textbf{(2)} sampling $n$ abstractions and then $n$ solutions per abstraction. In Table \ref{tab:passk_abs}, abstraction conditioning consistently outperforms pure solution sampling. For example, at $n=16$, abstraction conditioning achieves a pass@k of $0.71$ versus $0.65$ without abstractions, and at $n=256$ the gap widens to $0.87$ versus $0.82$. This corroborates the efficacy of abstractions, even under matched compute budgets.

\begin{figure}
    \centering
    \vspace{-0.3cm}
    \includegraphics[width=0.87\linewidth]{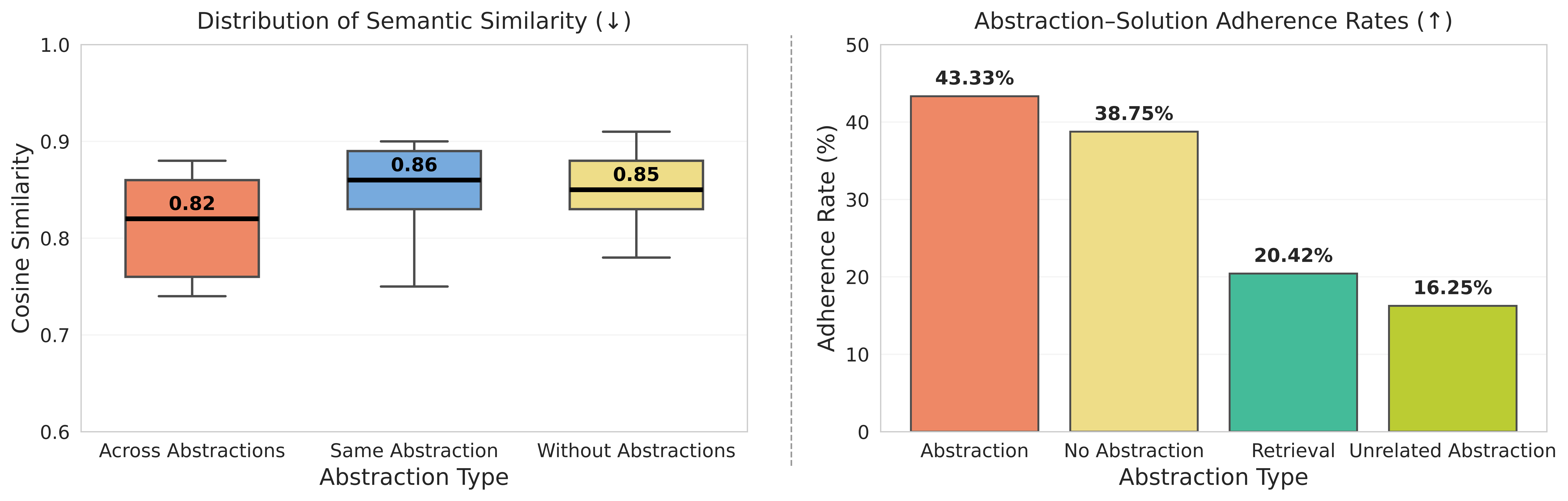}
    \vspace{-0.2cm}
      \caption{\footnotesize\textbf{Abstraction-conditioned solution generation analysis.} \methodname{} produces solutions with \textbf{(left)} greater semantic diversity across different abstractions and \textbf{(right)} higher abstraction adherence than baselines.}
    \label{fig:sol-analysis}
    \vspace{-0.4cm}
\end{figure}

\textcolor{lightblue}{\emph{\textbf{3) Understanding the behavior of the abstraction-conditioned solution generator.}}} A desirable property of the solution generator is the ability to follow the proposed abstractions. To study this, we prompt ({o4-mini}) to classify whether a particular solution trace produced by a trained solution generator closely adheres to a given abstraction. We ask for a binary decision on each pair of abstraction and solution, and measure the adherence rate across 200 such pairs. In Figure~\ref{fig:sol-analysis} (right), we report the adherence rates under four conditions: \textbf{(1)} \texttt{Abstraction}, where we measure adherence rates between an abstraction and a solution generated by conditioning on this abstraction itself; \textbf{(2)} \texttt{No abstraction}, where we measure the adherence rates between an abstraction and a solution generated by conditioning on the problem without any abstraction; \textbf{(3)} \texttt{Retrieval}, where we measure adherence rates between an abstraction and a semantically similar prior solution to the problem; and \textbf{(4)} \texttt{Unrelated abstraction}, where we measure the adherence rates between an abstraction and a solution generated via a different abstraction.
We find that the \texttt{Abstraction} condition achieves the highest adherence rate. Intuitively, this means that the trained solution generator is detected to be more likely to follow the strategy or guidance of the given abstraction it is conditioned on. 

Additionally, we measure the semantic similarity of two solutions generated without abstraction conditioning, two solutions generated by conditioning on the same abstraction, and two solutions generated by conditioning on different  abstractions in Figure~\ref{fig:sol-analysis} (left). We compute semantic similarity by encoding solutions into vector representations using Qwen3-4B and measuring the cosine similarity between normalized embeddings. We find that the semantic similarity between two solutions generated by conditioning on the same abstraction is highest, which is perhaps expected. Two solutions generated without conditioning on any abstraction are also quite similar semantically, whereas two solutions generated by conditioning on different abstractions are visibly less similar than the other two cases, implying that conditioning on and adhering to abstractions improves diversity of reasoning traces generated.

\begin{AIbox}{Takeaways: Experimental Results}
\begin{itemize}[itemsep=2pt]
\setlength{\leftskip}{-20pt}
    \item \methodname{} outperforms RL fine-tuning approaches that do not propose or leverage abstractions.
    \item Jointly scaling the number of abstractions and solution samples when deploying a model trained by \methodname{} enables continued performance gains even when scaling solutions alone saturates.
    \item Conditioning on various proposed abstractions produces solutions that are semantically more distinct from each other (Fig.~\ref{fig:sol-analysis} (left)) since \methodname{} adheres to the abstraction (Fig.~\ref{fig:sol-analysis} (right)).
\end{itemize} 
\end{AIbox}
\vspace{-0.2cm}
\section{Discussion and Perspectives on Future Work}
\vspace{-0.2cm}

We introduce reasoning abstractions: concise representations of procedural and factual knowledge that are expressed in natural language, as a means to broaden the reasoning strategies used by LLMs. Our method, \methodname{}, instantiates a two-player training framework that trains an abstraction generator and an abstraction-conditioned solution generator. \methodname{} yields consistent improvements across several mathematical reasoning benchmarks, outperforming existing methods for training LLMs to reason. Moreover, we show that allocating compute toward generating diverse abstractions, rather than increasing solution sampling alone, leads to greater performance gains. This highlights abstractions as a complementary axis for scaling test-time compute: while longer length chains-of-thoughts and parallel solution sampling provide some existing ways to scale compute, using abstractions provides us with an orthogonal axis to improve performance. This is particularly relevant as we start to see limited gains from standard scaling of length of chains of thought in reasoning. While we showed abstractions can be helpful, we limited our evaluation to mathematical reasoning tasks, leaving open‐ended reasoning unexplored. 

\textbf{Future work.} It is important to study \methodname{} can be extended to train a single model that can both propose and utilize abstractions. Perhaps this can be done by large-scale mid-training that showcases the approach of proposing high-level strategies followed by RL training akin to DeepSeek-R1~\citep{deepseekai2025deepseekr1incentivizingreasoningcapability}. We tried training a single model to do both abstraction generation and solution generation, after a lightweight SFT on traces showing questions paired with abstractions and corresponding solutions, but we found this approach to very quickly lose the ability of proposing abstractions over the course of RL training. We believe that this can be addressed via targeted mid-training or active interventions during RL, but this is an open direction that future work should study.
We also think that attempting to understand the phenomenon of how training with abstractions improves the performance without abstractions (Table~\ref{tab:main_results_correct}) is also intriguing. Likely this is a result of some form of generalization, but understanding the mechanisms behind this phenomena and how to amplify them is also interesting for future work.

\vspace{-0.2cm}
\section{Acknowledgements}
\vspace{-0.2cm}

We would like to thank Ian Wu, Matthew Yang, Rafael Rafailov, Yuejiang Liu, Violet Xiang, Joy He-Yueya, and others in the CMU AIRe lab and Stanford IRIS Lab for discussions and feedback. We thank Schmidt Sciences and the DSAI cluster, the FLAME cluster at CMU, and Delta for providing compute support for some of the critical experiments in this paper. We also thank Foundry for providing compute. AS gratefully acknowledges the support of the NSF Graduate Research Fellowship Program and Toyota Research Institute (TRI) for funding, compute, and API credits. CF was supported by Schmidt Sciences. This work was supported by the Schmidt Sciences AI2050 program, an Amazon gift, and the Office of Naval Research under N00014-24-12206. 
\clearpage

\bibliography{main}

\begin{thebibliography}{54}
\providecommand{\natexlab}[1]{#1}
\providecommand{\url}[1]{\texttt{#1}}
\expandafter\ifx\csname urlstyle\endcsname\relax
  \providecommand{\doi}[1]{doi: #1}\else
  \providecommand{\doi}{doi: \begingroup \urlstyle{rm}\Url}\fi

\bibitem[Alex et~al.(2021)Alex, Lifland, Tunstall, Thakur, Maham, Riedel, Hine, Ashurst, Sedille, Carlier, et~al.]{alex2021raft}
Neel Alex, Eli Lifland, Lewis Tunstall, Abhishek Thakur, Pegah Maham, C~Jess Riedel, Emmie Hine, Carolyn Ashurst, Paul Sedille, Alexis Carlier, et~al.
\newblock Raft: A real-world few-shot text classification benchmark.
\newblock \emph{arXiv preprint arXiv:2109.14076}, 2021.

\bibitem[Anonymous(2025)]{anonymous2025optimizing}
Anonymous.
\newblock Optimizing inference-time reasoning in {LLM}s via retrieval-augmented reflection, 2025.
\newblock URL \url{https://openreview.net/forum?id=ElYRG3pJcv}.

\bibitem[Borgeaud et~al.(2022)Borgeaud, Mensch, Hoffmann, Cai, Rutherford, Millican, Van Den~Driessche, Lespiau, Damoc, Clark, et~al.]{borgeaud2022improving}
Sebastian Borgeaud, Arthur Mensch, Jordan Hoffmann, Trevor Cai, Eliza Rutherford, Katie Millican, George~Bm Van Den~Driessche, Jean-Baptiste Lespiau, Bogdan Damoc, Aidan Clark, et~al.
\newblock Improving language models by retrieving from trillions of tokens.
\newblock In \emph{International conference on machine learning}, pages 2206--2240. PMLR, 2022.

\bibitem[Charniak and Johnson(2005)]{10.3115/1219840.1219862}
Eugene Charniak and Mark Johnson.
\newblock Coarse-to-fine n-best parsing and maxent discriminative reranking.
\newblock In \emph{Proceedings of the 43rd Annual Meeting on Association for Computational Linguistics}, ACL '05, page 173–180, USA, 2005. Association for Computational Linguistics.
\newblock \doi{10.3115/1219840.1219862}.
\newblock URL \url{https://doi.org/10.3115/1219840.1219862}.

\bibitem[Chen et~al.(2021)Chen, Tworek, Jun, Yuan, de~Oliveira~Pinto, Kaplan, Edwards, Burda, Joseph, Brockman, Ray, Puri, Krueger, Petrov, Khlaaf, Sastry, Mishkin, Chan, Gray, Ryder, Pavlov, Power, Kaiser, Bavarian, Winter, Tillet, Such, Cummings, Plappert, Chantzis, Barnes, Herbert-Voss, Guss, Nichol, Paino, Tezak, Tang, Babuschkin, Balaji, Jain, Saunders, Hesse, Carr, Leike, Achiam, Misra, Morikawa, Radford, Knight, Brundage, Murati, Mayer, Welinder, McGrew, Amodei, McCandlish, Sutskever, and Zaremba]{chen2021evaluatinglargelanguagemodels}
Mark Chen, Jerry Tworek, Heewoo Jun, Qiming Yuan, Henrique~Ponde de~Oliveira~Pinto, Jared Kaplan, Harri Edwards, Yuri Burda, Nicholas Joseph, Greg Brockman, Alex Ray, Raul Puri, Gretchen Krueger, Michael Petrov, Heidy Khlaaf, Girish Sastry, Pamela Mishkin, Brooke Chan, Scott Gray, Nick Ryder, Mikhail Pavlov, Alethea Power, Lukasz Kaiser, Mohammad Bavarian, Clemens Winter, Philippe Tillet, Felipe~Petroski Such, Dave Cummings, Matthias Plappert, Fotios Chantzis, Elizabeth Barnes, Ariel Herbert-Voss, William~Hebgen Guss, Alex Nichol, Alex Paino, Nikolas Tezak, Jie Tang, Igor Babuschkin, Suchir Balaji, Shantanu Jain, William Saunders, Christopher Hesse, Andrew~N. Carr, Jan Leike, Josh Achiam, Vedant Misra, Evan Morikawa, Alec Radford, Matthew Knight, Miles Brundage, Mira Murati, Katie Mayer, Peter Welinder, Bob McGrew, Dario Amodei, Sam McCandlish, Ilya Sutskever, and Wojciech Zaremba.
\newblock Evaluating large language models trained on code, 2021.
\newblock URL \url{https://arxiv.org/abs/2107.03374}.

\bibitem[DeepSeek-AI et~al.(2025)DeepSeek-AI, Guo, Yang, Zhang, Song, Zhang, Xu, Zhu, Ma, Wang, Bi, Zhang, Yu, Wu, Wu, Gou, Shao, Li, Gao, Liu, Xue, Wang, Wu, Feng, Lu, Zhao, Deng, Zhang, Ruan, Dai, Chen, Ji, Li, Lin, Dai, Luo, Hao, Chen, Li, Zhang, Bao, Xu, Wang, Ding, Xin, Gao, Qu, Li, Guo, Li, Wang, Chen, Yuan, Qiu, Li, Cai, Ni, Liang, Chen, Dong, Hu, Gao, Guan, Huang, Yu, Wang, Zhang, Zhao, Wang, Zhang, Xu, Xia, Zhang, Zhang, Tang, Li, Wang, Li, Tian, Huang, Zhang, Wang, Chen, Du, Ge, Zhang, Pan, Wang, Chen, Jin, Chen, Lu, Zhou, Chen, Ye, Wang, Yu, Zhou, Pan, Li, Zhou, Wu, Ye, Yun, Pei, Sun, Wang, Zeng, Zhao, Liu, Liang, Gao, Yu, Zhang, Xiao, An, Liu, Wang, Chen, Nie, Cheng, Liu, Xie, Liu, Yang, Li, Su, Lin, Li, Jin, Shen, Chen, Sun, Wang, Song, Zhou, Wang, Shan, Li, Wang, Wei, Zhang, Xu, Li, Zhao, Sun, Wang, Yu, Zhang, Shi, Xiong, He, Piao, Wang, Tan, Ma, Liu, Guo, Ou, Wang, Gong, Zou, He, Xiong, Luo, You, Liu, Zhou, Zhu, Xu, Huang, Li, Zheng, Zhu, Ma, Tang, Zha, Yan, Ren, Ren, Sha, Fu, Xu, Xie, Zhang,
  Hao, Ma, Yan, Wu, Gu, Zhu, Liu, Li, Xie, Song, Pan, Huang, Xu, Zhang, and Zhang]{deepseekai2025deepseekr1incentivizingreasoningcapability}
DeepSeek-AI, Daya Guo, Dejian Yang, Haowei Zhang, Junxiao Song, Ruoyu Zhang, Runxin Xu, Qihao Zhu, Shirong Ma, Peiyi Wang, Xiao Bi, Xiaokang Zhang, Xingkai Yu, Yu~Wu, Z.~F. Wu, Zhibin Gou, Zhihong Shao, Zhuoshu Li, Ziyi Gao, Aixin Liu, Bing Xue, Bingxuan Wang, Bochao Wu, Bei Feng, Chengda Lu, Chenggang Zhao, Chengqi Deng, Chenyu Zhang, Chong Ruan, Damai Dai, Deli Chen, Dongjie Ji, Erhang Li, Fangyun Lin, Fucong Dai, Fuli Luo, Guangbo Hao, Guanting Chen, Guowei Li, H.~Zhang, Han Bao, Hanwei Xu, Haocheng Wang, Honghui Ding, Huajian Xin, Huazuo Gao, Hui Qu, Hui Li, Jianzhong Guo, Jiashi Li, Jiawei Wang, Jingchang Chen, Jingyang Yuan, Junjie Qiu, Junlong Li, J.~L. Cai, Jiaqi Ni, Jian Liang, Jin Chen, Kai Dong, Kai Hu, Kaige Gao, Kang Guan, Kexin Huang, Kuai Yu, Lean Wang, Lecong Zhang, Liang Zhao, Litong Wang, Liyue Zhang, Lei Xu, Leyi Xia, Mingchuan Zhang, Minghua Zhang, Minghui Tang, Meng Li, Miaojun Wang, Mingming Li, Ning Tian, Panpan Huang, Peng Zhang, Qiancheng Wang, Qinyu Chen, Qiushi Du, Ruiqi Ge, Ruisong
  Zhang, Ruizhe Pan, Runji Wang, R.~J. Chen, R.~L. Jin, Ruyi Chen, Shanghao Lu, Shangyan Zhou, Shanhuang Chen, Shengfeng Ye, Shiyu Wang, Shuiping Yu, Shunfeng Zhou, Shuting Pan, S.~S. Li, Shuang Zhou, Shaoqing Wu, Shengfeng Ye, Tao Yun, Tian Pei, Tianyu Sun, T.~Wang, Wangding Zeng, Wanjia Zhao, Wen Liu, Wenfeng Liang, Wenjun Gao, Wenqin Yu, Wentao Zhang, W.~L. Xiao, Wei An, Xiaodong Liu, Xiaohan Wang, Xiaokang Chen, Xiaotao Nie, Xin Cheng, Xin Liu, Xin Xie, Xingchao Liu, Xinyu Yang, Xinyuan Li, Xuecheng Su, Xuheng Lin, X.~Q. Li, Xiangyue Jin, Xiaojin Shen, Xiaosha Chen, Xiaowen Sun, Xiaoxiang Wang, Xinnan Song, Xinyi Zhou, Xianzu Wang, Xinxia Shan, Y.~K. Li, Y.~Q. Wang, Y.~X. Wei, Yang Zhang, Yanhong Xu, Yao Li, Yao Zhao, Yaofeng Sun, Yaohui Wang, Yi~Yu, Yichao Zhang, Yifan Shi, Yiliang Xiong, Ying He, Yishi Piao, Yisong Wang, Yixuan Tan, Yiyang Ma, Yiyuan Liu, Yongqiang Guo, Yuan Ou, Yuduan Wang, Yue Gong, Yuheng Zou, Yujia He, Yunfan Xiong, Yuxiang Luo, Yuxiang You, Yuxuan Liu, Yuyang Zhou, Y.~X. Zhu,
  Yanhong Xu, Yanping Huang, Yaohui Li, Yi~Zheng, Yuchen Zhu, Yunxian Ma, Ying Tang, Yukun Zha, Yuting Yan, Z.~Z. Ren, Zehui Ren, Zhangli Sha, Zhe Fu, Zhean Xu, Zhenda Xie, Zhengyan Zhang, Zhewen Hao, Zhicheng Ma, Zhigang Yan, Zhiyu Wu, Zihui Gu, Zijia Zhu, Zijun Liu, Zilin Li, Ziwei Xie, Ziyang Song, Zizheng Pan, Zhen Huang, Zhipeng Xu, Zhongyu Zhang, and Zhen Zhang.
\newblock Deepseek-r1: Incentivizing reasoning capability in llms via reinforcement learning, 2025.
\newblock URL \url{https://arxiv.org/abs/2501.12948}.

\bibitem[Feng et~al.(2024)Feng, Wan, Wen, McAleer, Wen, Zhang, and Wang]{feng2024alphazeroliketreesearchguidelarge}
Xidong Feng, Ziyu Wan, Muning Wen, Stephen~Marcus McAleer, Ying Wen, Weinan Zhang, and Jun Wang.
\newblock Alphazero-like tree-search can guide large language model decoding and training, 2024.
\newblock URL \url{https://arxiv.org/abs/2309.17179}.

\bibitem[Fernando et~al.(2023)Fernando, Banarse, Michalewski, Osindero, and Rockt{\"a}schel]{fernando2023promptbreeder}
Chrisantha Fernando, Dylan Banarse, Henryk Michalewski, Simon Osindero, and Tim Rockt{\"a}schel.
\newblock Promptbreeder: Self-referential self-improvement via prompt evolution.
\newblock \emph{arXiv preprint arXiv:2309.16797}, 2023.

\bibitem[Gou et~al.(2023)Gou, Shao, Gong, Shen, Yang, Duan, and Chen]{gou2023critic}
Zhibin Gou, Zhihong Shao, Yeyun Gong, Yelong Shen, Yujiu Yang, Nan Duan, and Weizhu Chen.
\newblock Critic: Large language models can self-correct with tool-interactive critiquing.
\newblock \emph{arXiv preprint arXiv:2305.11738}, 2023.

\bibitem[Guha et~al.(2023)Guha, Nyarko, Ho, R{\'e}, Chilton, Chohlas-Wood, Peters, Waldon, Rockmore, Zambrano, et~al.]{guha2023legalbench}
Neel Guha, Julian Nyarko, Daniel Ho, Christopher R{\'e}, Adam Chilton, Alex Chohlas-Wood, Austin Peters, Brandon Waldon, Daniel Rockmore, Diego Zambrano, et~al.
\newblock Legalbench: A collaboratively built benchmark for measuring legal reasoning in large language models.
\newblock \emph{Advances in Neural Information Processing Systems}, 36:\penalty0 44123--44279, 2023.

\bibitem[Hao et~al.(2023)Hao, Gu, Ma, Hong, Wang, Wang, and Hu]{hao2023reasoninglanguagemodelplanning}
Shibo Hao, Yi~Gu, Haodi Ma, Joshua~Jiahua Hong, Zhen Wang, Daisy~Zhe Wang, and Zhiting Hu.
\newblock Reasoning with language model is planning with world model, 2023.
\newblock URL \url{https://arxiv.org/abs/2305.14992}.

\bibitem[Ho et~al.(2023)Ho, Schmid, and Yun]{ho2023largelanguagemodelsreasoning}
Namgyu Ho, Laura Schmid, and Se-Young Yun.
\newblock Large language models are reasoning teachers, 2023.
\newblock URL \url{https://arxiv.org/abs/2212.10071}.

\bibitem[Kumar et~al.(2024)Kumar, Zhuang, Agarwal, Su, Co-Reyes, Singh, Baumli, Iqbal, Bishop, Roelofs, Zhang, McKinney, Shrivastava, Paduraru, Tucker, Precup, Behbahani, and Faust]{kumar2024traininglanguagemodelsselfcorrect}
Aviral Kumar, Vincent Zhuang, Rishabh Agarwal, Yi~Su, John~D Co-Reyes, Avi Singh, Kate Baumli, Shariq Iqbal, Colton Bishop, Rebecca Roelofs, Lei~M Zhang, Kay McKinney, Disha Shrivastava, Cosmin Paduraru, George Tucker, Doina Precup, Feryal Behbahani, and Aleksandra Faust.
\newblock Training language models to self-correct via reinforcement learning, 2024.
\newblock URL \url{https://arxiv.org/abs/2409.12917}.

\bibitem[Lewis et~al.(2020)Lewis, Perez, Piktus, Petroni, Karpukhin, Goyal, K{\"u}ttler, Lewis, Yih, Rockt{\"a}schel, et~al.]{lewis2020retrieval}
Patrick Lewis, Ethan Perez, Aleksandra Piktus, Fabio Petroni, Vladimir Karpukhin, Naman Goyal, Heinrich K{\"u}ttler, Mike Lewis, Wen-tau Yih, Tim Rockt{\"a}schel, et~al.
\newblock Retrieval-augmented generation for knowledge-intensive nlp tasks.
\newblock \emph{Advances in neural information processing systems}, 33:\penalty0 9459--9474, 2020.

\bibitem[Li et~al.(2023)Li, Hammoud, Itani, Khizbullin, and Ghanem]{li2023camelcommunicativeagentsmind}
Guohao Li, Hasan Abed Al~Kader Hammoud, Hani Itani, Dmitrii Khizbullin, and Bernard Ghanem.
\newblock Camel: Communicative agents for "mind" exploration of large language model society, 2023.
\newblock URL \url{https://arxiv.org/abs/2303.17760}.

\bibitem[Li et~al.(2024)Li, Hu, Larsen, Wu, Alford, Woo, Dunn, Tang, Naim, Nguyen, Zheng, Tavares, Pu, and Ellis]{li2024combininginductiontransductionabstract}
Wen-Ding Li, Keya Hu, Carter Larsen, Yuqing Wu, Simon Alford, Caleb Woo, Spencer~M. Dunn, Hao Tang, Michelangelo Naim, Dat Nguyen, Wei-Long Zheng, Zenna Tavares, Yewen Pu, and Kevin Ellis.
\newblock Combining induction and transduction for abstract reasoning, 2024.
\newblock URL \url{https://arxiv.org/abs/2411.02272}.

\bibitem[Li et~al.(2025)Li, Dong, Jin, Zhang, Zhou, Zhu, Zhang, and Dou]{li2025searcho1agenticsearchenhancedlarge}
Xiaoxi Li, Guanting Dong, Jiajie Jin, Yuyao Zhang, Yujia Zhou, Yutao Zhu, Peitian Zhang, and Zhicheng Dou.
\newblock Search-o1: Agentic search-enhanced large reasoning models, 2025.
\newblock URL \url{https://arxiv.org/abs/2501.05366}.

\bibitem[Lin et~al.(2025)Lin, Snell, Wang, Packer, Wooders, Stoica, and Gonzalez]{lin2025sleep}
Kevin Lin, Charlie Snell, Yu~Wang, Charles Packer, Sarah Wooders, Ion Stoica, and Joseph~E Gonzalez.
\newblock Sleep-time compute: Beyond inference scaling at test-time.
\newblock \emph{arXiv preprint arXiv:2504.13171}, 2025.

\bibitem[Luo et~al.(2025)Luo, Tan, Wong, Shi, Tang, Roongta, Cai, Luo, Zhang, Li, Popa, and Stoica]{deepscaler2025}
Michael Luo, Sijun Tan, Justin Wong, Xiaoxiang Shi, William Tang, Manan Roongta, Colin Cai, Jeffrey Luo, Tianjun Zhang, Erran Li, Raluca~Ada Popa, and Ion Stoica.
\newblock Deepscaler: Surpassing o1-preview with a 1.5b model by scaling rl, 2025.
\newblock Notion Blog.

\bibitem[{Madaan} et~al.(2023){Madaan}, {Tandon}, {Gupta}, {Hallinan}, {Gao}, {Wiegreffe}, {Alon}, {Dziri}, {Prabhumoye}, {Yang}, {Gupta}, {Prasad Majumder}, {Hermann}, {Welleck}, {Yazdanbakhsh}, and {Clark}]{2023arXiv230317651M}
Aman {Madaan}, Niket {Tandon}, Prakhar {Gupta}, Skyler {Hallinan}, Luyu {Gao}, Sarah {Wiegreffe}, Uri {Alon}, Nouha {Dziri}, Shrimai {Prabhumoye}, Yiming {Yang}, Shashank {Gupta}, Bodhisattwa {Prasad Majumder}, Katherine {Hermann}, Sean {Welleck}, Amir {Yazdanbakhsh}, and Peter {Clark}.
\newblock {Self-Refine: Iterative Refinement with Self-Feedback}.
\newblock \emph{arXiv e-prints}, art. arXiv:2303.17651, March 2023.
\newblock \doi{10.48550/arXiv.2303.17651}.

\bibitem[Madaan et~al.(2023)Madaan, Tandon, Gupta, Hallinan, Gao, Wiegreffe, Alon, Dziri, Prabhumoye, Yang, et~al.]{madaan2023self}
Aman Madaan, Niket Tandon, Prakhar Gupta, Skyler Hallinan, Luyu Gao, Sarah Wiegreffe, Uri Alon, Nouha Dziri, Shrimai Prabhumoye, Yiming Yang, et~al.
\newblock Self-refine: Iterative refinement with self-feedback.
\newblock \emph{Advances in Neural Information Processing Systems}, 36:\penalty0 46534--46594, 2023.

\bibitem[{Mathematical Association of America}(2025)]{AIME2025}
{Mathematical Association of America}.
\newblock {2025 American Invitational Mathematics Examination (AIME) I: Problems and Solutions}, February 2025.
\newblock URL \url{https://artofproblemsolving.com/wiki/index.php/2025_AIME_I}.
\newblock Art of Problem Solving Wiki entry.

\bibitem[Menon et~al.(2022)Menon, Ghosh, and Srivastava]{menon2022clues}
Rakesh~R. Menon, Sayan Ghosh, and Shashank Srivastava.
\newblock Clues a benchmark for learning classifiers using natural language explanations.
\newblock 2022.

\bibitem[Nye et~al.(2021)Nye, Andreassen, Gur-Ari, Michalewski, Austin, Bieber, Dohan, Lewkowycz, Bosma, Luan, Sutton, and Odena]{nye2021workscratchpadsintermediatecomputation}
Maxwell Nye, Anders~Johan Andreassen, Guy Gur-Ari, Henryk Michalewski, Jacob Austin, David Bieber, David Dohan, Aitor Lewkowycz, Maarten Bosma, David Luan, Charles Sutton, and Augustus Odena.
\newblock Show your work: Scratchpads for intermediate computation with language models, 2021.
\newblock URL \url{https://arxiv.org/abs/2112.00114}.

\bibitem[Pan et~al.(2025)Pan, Li, Lian, Snell, Zhou, Yala, Darrell, Keutzer, and Suhr]{pan2025learning}
Jiayi Pan, Xiuyu Li, Long Lian, Charlie Snell, Yifei Zhou, Adam Yala, Trevor Darrell, Kurt Keutzer, and Alane Suhr.
\newblock Learning adaptive parallel reasoning with language models.
\newblock \emph{arXiv preprint arXiv:2504.15466}, 2025.

\bibitem[Pryzant et~al.(2023)Pryzant, Iter, Li, Lee, Zhu, and Zeng]{pryzant2023automatic}
Reid Pryzant, Dan Iter, Jerry Li, Yin~Tat Lee, Chenguang Zhu, and Michael Zeng.
\newblock Automatic prompt optimization with" gradient descent" and beam search.
\newblock \emph{arXiv preprint arXiv:2305.03495}, 2023.

\bibitem[{Qu} et~al.(2024){Qu}, {Zhang}, {Garg}, and {Kumar}]{2024arXiv240718219Q}
Yuxiao {Qu}, Tianjun {Zhang}, Naman {Garg}, and Aviral {Kumar}.
\newblock {Recursive Introspection: Teaching Language Model Agents How to Self-Improve}.
\newblock \emph{arXiv e-prints}, art. arXiv:2407.18219, July 2024.
\newblock \doi{10.48550/arXiv.2407.18219}.

\bibitem[Qu et~al.(2024)Qu, Zhang, Garg, and Kumar]{qu2024recursive}
Yuxiao Qu, Tianjun Zhang, Naman Garg, and Aviral Kumar.
\newblock Recursive introspection: Teaching language model agents how to self-improve.
\newblock \emph{arXiv preprint arXiv:2407.18219}, 2024.

\bibitem[Qu et~al.(2025)Qu, Yang, Setlur, Tunstall, Beeching, Salakhutdinov, and Kumar]{qu2025optimizing}
Yuxiao Qu, Matthew~YR Yang, Amrith Setlur, Lewis Tunstall, Edward~Emanuel Beeching, Ruslan Salakhutdinov, and Aviral Kumar.
\newblock Optimizing test-time compute via meta reinforcement fine-tuning.
\newblock \emph{arXiv preprint arXiv:2503.07572}, 2025.

\bibitem[{Qwen Team}(2025)]{qwen3_2025}
{Qwen Team}.
\newblock Qwen3 technical report.
\newblock Technical report, Qwen, May 2025.
\newblock URL \url{https://huggingface.co/Qwen}.
\newblock Available at: https://huggingface.co/Qwen, https://modelscope.cn/organization/qwen, https://github.com/QwenLM/Qwen3.

\bibitem[Rafailov et~al.(2023)Rafailov, Sharma, Mitchell, Manning, Ermon, and Finn]{rafailov2023direct}
Rafael Rafailov, Archit Sharma, Eric Mitchell, Christopher~D Manning, Stefano Ermon, and Chelsea Finn.
\newblock Direct preference optimization: Your language model is secretly a reward model.
\newblock \emph{Advances in Neural Information Processing Systems}, 36:\penalty0 53728--53741, 2023.

\bibitem[Schäfer et~al.(2020)Schäfer, Ohme, and Nitz]{Sch_fer_2020}
Marlin~B. Schäfer, Frank Ohme, and Alexander~H. Nitz.
\newblock Detection of gravitational-wave signals from binary neutron star mergers using machine learning.
\newblock \emph{Physical Review D}, 102\penalty0 (6), September 2020.
\newblock ISSN 2470-0029.
\newblock \doi{10.1103/physrevd.102.063015}.
\newblock URL \url{http://dx.doi.org/10.1103/PhysRevD.102.063015}.

\bibitem[Setlur et~al.(2025)Setlur, Yang, Snell, Greer, Wu, Smith, Simchowitz, and Kumar]{e3_extrapolation_2025}
Amrith Setlur, Matthew Y.~R. Yang, Charlie Snell, Jeremy Greer, Ian Wu, Virginia Smith, Max Simchowitz, and Aviral Kumar.
\newblock e3: Learning to explore enables extrapolation of test-time compute for llms.
\newblock 2025.
\newblock URL \url{https://arxiv.org/abs/2506.09026}.

\bibitem[Shao et~al.(2024)Shao, Wang, Zhu, Xu, Song, Bi, Zhang, Zhang, Li, Wu, et~al.]{shao2024deepseekmath}
Zhihong Shao, Peiyi Wang, Qihao Zhu, Runxin Xu, Junxiao Song, Xiao Bi, Haowei Zhang, Mingchuan Zhang, YK~Li, Y~Wu, et~al.
\newblock Deepseekmath: Pushing the limits of mathematical reasoning in open language models.
\newblock \emph{arXiv preprint arXiv:2402.03300}, 2024.

\bibitem[Shinn et~al.(2023)Shinn, Cassano, Gopinath, Narasimhan, and Yao]{shinn2023reflexion}
Noah Shinn, Federico Cassano, Ashwin Gopinath, Karthik Narasimhan, and Shunyu Yao.
\newblock Reflexion: Language agents with verbal reinforcement learning.
\newblock \emph{Advances in Neural Information Processing Systems}, 36:\penalty0 8634--8652, 2023.

\bibitem[{Snell} et~al.(2024){Snell}, {Lee}, {Xu}, and {Kumar}]{2024arXiv240803314S}
Charlie {Snell}, Jaehoon {Lee}, Kelvin {Xu}, and Aviral {Kumar}.
\newblock {Scaling LLM Test-Time Compute Optimally can be More Effective than Scaling Model Parameters}.
\newblock \emph{arXiv e-prints}, art. arXiv:2408.03314, August 2024.
\newblock \doi{10.48550/arXiv.2408.03314}.

\bibitem[Snell et~al.(2024)Snell, Lee, Xu, and Kumar]{snell2024scaling}
Charlie Snell, Jaehoon Lee, Kelvin Xu, and Aviral Kumar.
\newblock Scaling llm test-time compute optimally can be more effective than scaling model parameters.
\newblock \emph{arXiv preprint arXiv:2408.03314}, 2024.

\bibitem[Suzgun et~al.(2025)Suzgun, Yuksekgonul, Bianchi, Jurafsky, and Zou]{suzgun2025dynamic}
Mirac Suzgun, Mert Yuksekgonul, Federico Bianchi, Dan Jurafsky, and James Zou.
\newblock Dynamic cheatsheet: Test-time learning with adaptive memory.
\newblock \emph{arXiv preprint arXiv:2504.07952}, 2025.

\bibitem[Trivedi et~al.(2022)Trivedi, Balasubramanian, Khot, and Sabharwal]{trivedi2022interleaving}
Harsh Trivedi, Niranjan Balasubramanian, Tushar Khot, and Ashish Sabharwal.
\newblock Interleaving retrieval with chain-of-thought reasoning for knowledge-intensive multi-step questions.
\newblock \emph{arXiv preprint arXiv:2212.10509}, 2022.

\bibitem[Uesato et~al.(2022)Uesato, Kushman, Kumar, Song, Siegel, Wang, Creswell, Irving, and Higgins]{uesato2022solvingmathwordproblems}
Jonathan Uesato, Nate Kushman, Ramana Kumar, Francis Song, Noah Siegel, Lisa Wang, Antonia Creswell, Geoffrey Irving, and Irina Higgins.
\newblock Solving math word problems with process- and outcome-based feedback, 2022.
\newblock URL \url{https://arxiv.org/abs/2211.14275}.

\bibitem[{Verma} et~al.(2024){Verma}, {Midigeshi}, {Sinha}, {Solin}, {Natarajan}, and {Sharma}]{2024arXiv241020753V}
Prakhar {Verma}, Sukruta~Prakash {Midigeshi}, Gaurav {Sinha}, Arno {Solin}, Nagarajan {Natarajan}, and Amit {Sharma}.
\newblock {Plan$\times$RAG: Planning-guided Retrieval Augmented Generation}.
\newblock \emph{arXiv e-prints}, art. arXiv:2410.20753, October 2024.
\newblock \doi{10.48550/arXiv.2410.20753}.

\bibitem[Wang et~al.(2023)Wang, Wei, Schuurmans, Le, Chi, Narang, Chowdhery, and Zhou]{wang2023selfconsistencyimproveschainthought}
Xuezhi Wang, Jason Wei, Dale Schuurmans, Quoc Le, Ed~Chi, Sharan Narang, Aakanksha Chowdhery, and Denny Zhou.
\newblock Self-consistency improves chain of thought reasoning in language models, 2023.
\newblock URL \url{https://arxiv.org/abs/2203.11171}.

\bibitem[Wang et~al.(2025)Wang, Liu, Xu, Liang, Chen, He, Song, Yu, Li, Zhang, et~al.]{wang2025thoughts}
Yue Wang, Qiuzhi Liu, Jiahao Xu, Tian Liang, Xingyu Chen, Zhiwei He, Linfeng Song, Dian Yu, Juntao Li, Zhuosheng Zhang, et~al.
\newblock Thoughts are all over the place: On the underthinking of o1-like llms.
\newblock \emph{arXiv preprint arXiv:2501.18585}, 2025.

\bibitem[Yang et~al.(2023)Yang, Wang, Lu, Liu, Le, Zhou, and Chen]{yang2023large}
Chengrun Yang, Xuezhi Wang, Yifeng Lu, Hanxiao Liu, Quoc~V Le, Denny Zhou, and Xinyun Chen.
\newblock Large language models as optimizers.
\newblock \emph{arXiv preprint arXiv:2309.03409}, 2023.

\bibitem[Yao et~al.(2023{\natexlab{a}})Yao, Yu, Zhao, Shafran, Griffiths, Cao, and Narasimhan]{yao2023treethoughtsdeliberateproblem}
Shunyu Yao, Dian Yu, Jeffrey Zhao, Izhak Shafran, Thomas~L. Griffiths, Yuan Cao, and Karthik Narasimhan.
\newblock Tree of thoughts: Deliberate problem solving with large language models, 2023{\natexlab{a}}.
\newblock URL \url{https://arxiv.org/abs/2305.10601}.

\bibitem[Yao et~al.(2023{\natexlab{b}})Yao, Zhao, Yu, Du, Shafran, Narasimhan, and Cao]{yao2023reactsynergizingreasoningacting}
Shunyu Yao, Jeffrey Zhao, Dian Yu, Nan Du, Izhak Shafran, Karthik Narasimhan, and Yuan Cao.
\newblock React: Synergizing reasoning and acting in language models, 2023{\natexlab{b}}.
\newblock URL \url{https://arxiv.org/abs/2210.03629}.

\bibitem[Yu et~al.(2025)Yu, Zhang, Zhu, Yuan, Zuo, Yue, Fan, Liu, Liu, Liu, et~al.]{yu2025dapo}
Qiying Yu, Zheng Zhang, Ruofei Zhu, Yufeng Yuan, Xiaochen Zuo, Yu~Yue, Tiantian Fan, Gaohong Liu, Lingjun Liu, Xin Liu, et~al.
\newblock Dapo: An open-source llm reinforcement learning system at scale.
\newblock \emph{arXiv preprint arXiv:2503.14476}, 2025.

\bibitem[Yuan et~al.(2023)Yuan, Yuan, Li, Dong, Tan, and Zhou]{yuan2023scaling}
Zheng Yuan, Hongyi Yuan, Chengpeng Li, Guanting Dong, Chuanqi Tan, and Chang Zhou.
\newblock Scaling relationship on learning mathematical reasoning with large language models.
\newblock \emph{arXiv preprint arXiv:2308.01825}, 2023.

\bibitem[Yue et~al.(2025)Yue, Chen, Lu, Zhao, Wang, Song, and Huang]{yue2025does}
Yang Yue, Zhiqi Chen, Rui Lu, Andrew Zhao, Zhaokai Wang, Shiji Song, and Gao Huang.
\newblock Does reinforcement learning really incentivize reasoning capacity in llms beyond the base model?
\newblock \emph{arXiv preprint arXiv:2504.13837}, 2025.

\bibitem[Yuksekgonul et~al.(2025)Yuksekgonul, Bianchi, Boen, Liu, Lu, Huang, Guestrin, and Zou]{yuksekgonul2025optimizing}
Mert Yuksekgonul, Federico Bianchi, Joseph Boen, Sheng Liu, Pan Lu, Zhi Huang, Carlos Guestrin, and James Zou.
\newblock Optimizing generative ai by backpropagating language model feedback.
\newblock \emph{Nature}, 639\penalty0 (8055):\penalty0 609--616, 2025.

\bibitem[Zelikman et~al.(2022{\natexlab{a}})Zelikman, Wu, Mu, and Goodman]{zelikman2022star}
Eric Zelikman, Yuhuai Wu, Jesse Mu, and Noah Goodman.
\newblock Star: Bootstrapping reasoning with reasoning.
\newblock \emph{Advances in Neural Information Processing Systems}, 35:\penalty0 15476--15488, 2022{\natexlab{a}}.

\bibitem[Zelikman et~al.(2022{\natexlab{b}})Zelikman, Wu, Mu, and Goodman]{zelikman2022starbootstrappingreasoningreasoning}
Eric Zelikman, Yuhuai Wu, Jesse Mu, and Noah~D. Goodman.
\newblock Star: Bootstrapping reasoning with reasoning, 2022{\natexlab{b}}.
\newblock URL \url{https://arxiv.org/abs/2203.14465}.

\bibitem[Zhao et~al.(2024)Zhao, Huang, Xu, Lin, Liu, and Huang]{zhao2024expel}
Andrew Zhao, Daniel Huang, Quentin Xu, Matthieu Lin, Yong-Jin Liu, and Gao Huang.
\newblock Expel: Llm agents are experiential learners.
\newblock In \emph{Proceedings of the AAAI Conference on Artificial Intelligence}, volume~38, pages 19632--19642, 2024.

\bibitem[Zhou et~al.(2022)Zhou, Muresanu, Han, Paster, Pitis, Chan, and Ba]{zhou2022large}
Yongchao Zhou, Andrei~Ioan Muresanu, Ziwen Han, Keiran Paster, Silviu Pitis, Harris Chan, and Jimmy Ba.
\newblock Large language models are human-level prompt engineers.
\newblock In \emph{The Eleventh International Conference on Learning Representations}, 2022.

\end{thebibliography}

\newpage
\appendix
\onecolumn

\newpage 

\part*{Appendices}

\section{Experimental Details}
\vspace{-0.2cm}
\subsection{Pseudocode for \methodname{}}
\label{app:exp_details}
\vspace{-0.2cm}
\begin{algorithm}[ht]
\caption{Joint RL Training of $\absgen$ and $\solgen$}
\label{alg:abstraction}
\begin{algorithmic}[1]
\small
\Require 
  Policies $\absgen(\bz\mid\bx)$, $\solgen(\tilde{\by}\mid\bx,\bz)$  
  Datasets $\mathcal{D}_{\absgen},\;\mathcal{D}_{\solgen}$; rates $\alpha_{\absgen},\alpha_{\solgen}$; batch sizes $N,M$; epochs $E$
\State Initialize $\absgen,\solgen$
\For{$e=1$ to $E$}
  \Comment{Update abstraction policy}
  \For{$\{\bx_i\}_{i=1}^N\!\sim\!\mathcal{D}_{\absgen}$}
    \State $\bz_i\sim\pi^\mathrm{abs}_\theta(\cdot|\bx_i)$
    \State $r_i\leftarrow r_{\solgen}(\bx_i,\bz_i)$
    \State $\absgen\gets\absgen-\alpha_{\absgen}\nabla_{\absgen}\mathcal{L}_{\mathrm{STAR/RPO}}(\absgen;\bx_i,\bz_i,r_i)$
  \EndFor
  \Comment{Update solution policy}
  \For{$\{\bx_j\}_{j=1}^M\!\sim\!\mathcal{D}_{\solgen}$}
    \State $\bz_j\sim\absgen(\cdot|\bx_j)$, \quad $\tilde{\by}_j\sim\solgen(\cdot|\bx_j,\bz_j)$
    \State $r_j\leftarrow r(\bx_j,\bz_j,\tilde{\by}_j)$
    \State $\solgen\gets\solgen-\alpha_{\solgen}\nabla_{\solgen}\mathcal{L}_{\mathrm{GRPO}}(\solgen;\bx_j,\bz_j,\tilde{\by}_j,r_j)$
  \EndFor
\EndFor
\end{algorithmic}
\end{algorithm}

\subsection{Hyperparameters}
\begin{table*}[h]
\centering
\begin{tabularx}{0.55\linewidth}{l|c}
  \toprule
\multicolumn{1}{c}{\textbf{Hyperparameter}} \vline & \multicolumn{1}{c}{\textbf{Value}} \\
\midrule
\small
algorithm                             & DaPO~\citep{yu2025dapo} \\
training steps                        & 100 \\
epochs                                & 10 \\
train batch size                      & 128 \\
max prompt length                     & 3072 \\
max response length                   & 16384 \\
max extrapolation length              & 32768 \\
learning rate                         & 1e-6 \\
clip ratio (low / high)              & 0.2 / 0.5 \\
entropy coefficient                   & 0.001 \\
KL loss coefficient                   & 0.001 \\
KL loss type                          & low variance kl \\
sampling temperature (train / val)   & 0.6 / 0.6 \\
samples per prompt (train / val)     & 16 / 8 \\
max batched tokens                    & 32768 \\
\bottomrule
\end{tabularx}
\caption{Key training hyperparameters used in \methodname{}.}
\label{tab:ppo_hyper}
\end{table*}

\section{Additional Experimental Results}
\label{app:add_result}
\subsection{\methodname{}'s w/ abs performance on AIME 2024 and HMMT 2025}
 In this section, we evaluate the performance of the base model (Qwen-3-1.7B), GRPO-enhanced model, and our proposed method \methodname{} on two math reasoning benchmarks: AIME 2024 and HMMT 2025. As shown in Table~\ref{tab:main_results}, our method achieves the best performance across both datasets.

It is important to note that \methodname{} is trained using access to abstractions, yet it also generalizes better even when evaluated without abstraction. This suggests that \methodname{} does not merely overfit to the abstraction format but instead learns to effectively leverage high-level procedural guidance, leading to better generalization on challenging reasoning benchmarks.

\label{sec:AIME24_HMMT25}
\begin{table*}[ht]
  \centering
  \footnotesize
  \def\arraystretch{.99}
  \setlength{\tabcolsep}{0.42em}
  \begin{tabularx}{0.43\linewidth}{l|c|c}
    \toprule
    \multirow{1}{*}{\textbf{Approach}} & 
    \multicolumn{1}{c|}{\textbf{AIME 2024}} & 
    \multicolumn{1}{c}{\textbf{HMMT 2025}} \\ 
    \midrule
    Qwen-3-1.7B     & 48.54 & 22.50 \\
    + DaPO          & 44.17 & 23.13\\
    + \methodname{} & \textbf{51.46} & \textbf{23.75}\\
    \bottomrule
  \end{tabularx}
  \vspace{-0.1cm}
  \caption{\footnotesize{\textbf{\methodname{}'s w/ abs performance on AIME 2024 and HMMT 2025.}} \methodname{} outperforms both DaPO and the base Qwen model on the AIME 2024 and HMMT 2025 benchmarks.}
  \label{tab:main_results}
  \vspace{-0.4cm}
\end{table*}

\subsection{Design Choice Ablations}
\label{sec:design_choice_ablation}

In this section, we conduct ablation studies to isolate the contributions of key components in \methodname{}. We investigate three primary design choices, with results summarized in Table~\ref{tab:design_choice_ablation}: \textbf{(a)} utilizing a curriculum training strategy, \textbf{(b)} including prompts not annotated with an abstraction, and \textbf{(c)} applying reward masking to those non-annotated prompts.

\textbf{Curriculum training} refers to a staged process where the model first learns from simpler problems and gradually transitions to harder ones. We use the protocol from \citet{e3_extrapolation_2025} as inspiration, who demonstrated its effectiveness for direct math problem-solving. In our setting, which incorporates abstractions, curriculum training also proves beneficial, improving both average and best-case performance from 0.38 and 0.43 to 0.41 and 0.48, respectively, compared to non-curriculum training.

Next, we analyze the practice of \textbf{including prompts without abstractions} and applying \textbf{reward masking}. Including a small fraction of these "no-abstraction" prompts is intended to better condition the solution-generator on the abstractions when they are present. However, this risks the model learning a shortcut by simply ignoring the abstractions. To mitigate this, we apply reward masking: for completions on no-abstraction prompts, we nullify the policy reward by zeroing out the advantage, while retaining the KL penalty for regularization. This prevents the model from over-optimizing on examples that lack abstractions, a behavior that would otherwise hinder generalization.

Our findings confirm the efficacy of this combined approach. As shown in Table~\ref{tab:design_choice_ablation}, including no-abstraction prompts with reward masking is critical for performance. Ultimately, the combination of all three design choices—curriculum training, the inclusion of no-abstraction prompts, and reward masking—significantly outperforms alternative configurations.

\definecolor{darkgreen}{RGB}{0,150,0}
\newcommand{\cmark}{\textcolor{darkgreen}{\ding{51}}}
\newcommand{\xmark}{\textcolor{red}{\ding{55}}}
\begin{table*}[htbp]
  \centering
  \footnotesize
  \def\arraystretch{.99}
  \setlength{\tabcolsep}{0.42em}
  \resizebox{0.9\linewidth}{!}{
  \begin{tabularx}{1.05\linewidth}{l|c|c|c|cc}
    \toprule
    \multirow{2}{*}{\textbf{Approach}} & 
    \multicolumn{3}{c|}{\textbf{Design Choice}} & 
    \multicolumn{2}{c}{\textbf{AIME 2025}} \\ 
    & curriculum training & including no-abstraction prompt & reward masking & w/ abs (avg) & w/ abs (best) \\
    \midrule
    variant 1    & \xmark & \cmark & \xmark & 36.51 & 42.29 \\
    variant 2    & \xmark & \xmark & - & 37.08 & 42.50 \\
    variant 3    & \xmark & \cmark & \cmark & 37.50 & 43.33 \\
    \methodname{} & \cmark & \cmark & \cmark & \textbf{42.45} & \textbf{48.33} \\
    \bottomrule
  \end{tabularx}
  }
  \vspace{-0.1cm}
  \caption{\footnotesize{\textbf{Design Choices in \methodname{}.} We isolate the effects of curriculum training, no-abstraction inclusion, and reward masking. The full method achieves the strongest performance under abstraction-conditioned evaluation.}}
  \label{tab:design_choice_ablation}
  \vspace{-0.4cm}
\end{table*}

\subsection{Full Results for Abstractions in non-math domains}
\label{sec:full_details_nonmath}
As seen in Table~\ref{tab:diverse_domains}, conditioning on abstractions helps for 37 domains, where the average and best abstractions outperform standard prompting by 18.0\% and 30.0\% on average, respectively.

For ARC-AGI, we warmstart the abstraction generator model with synthetically augmented human annotations from the BARC dataset~\citep{li2024combininginductiontransductionabstract}. We present experiments with a fine-tuned abstraction generator and a frozen solution generator in Table 1. 

\section{Abstractions in Other Domains}
\label{app:abstractions_in_other_domains}

\begin{figure}[h]
\centering
\includegraphics[width=0.99\linewidth]{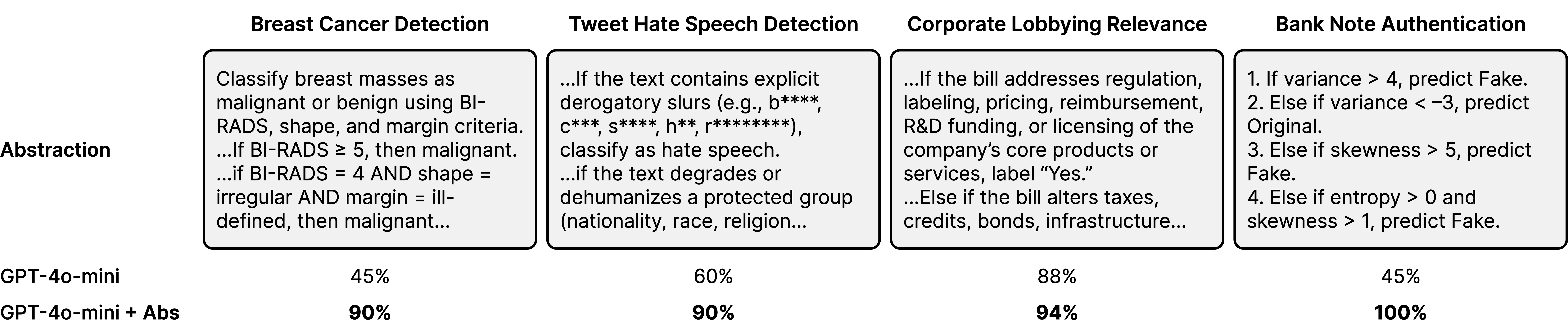}
\vspace{-1mm}
\caption{\footnotesize{\textbf{\emph{Examples of good reasoning abstractions in non-math domains}}. Adding the abstraction to the prompt of GPT-4o-mini consistently improves performance on unseen instances.}}
\label{fig:abstraction_examples}
\vspace{-0.2cm}
\end{figure}

\section{Qualitative Examples of Math Reasoning Abstractions}
\label{app:abstractions_examples}

\begin{wrapfigure}{r}{0.5\linewidth}
  \vspace{-4mm}
  \centering
  \includegraphics[width=0.8\linewidth]{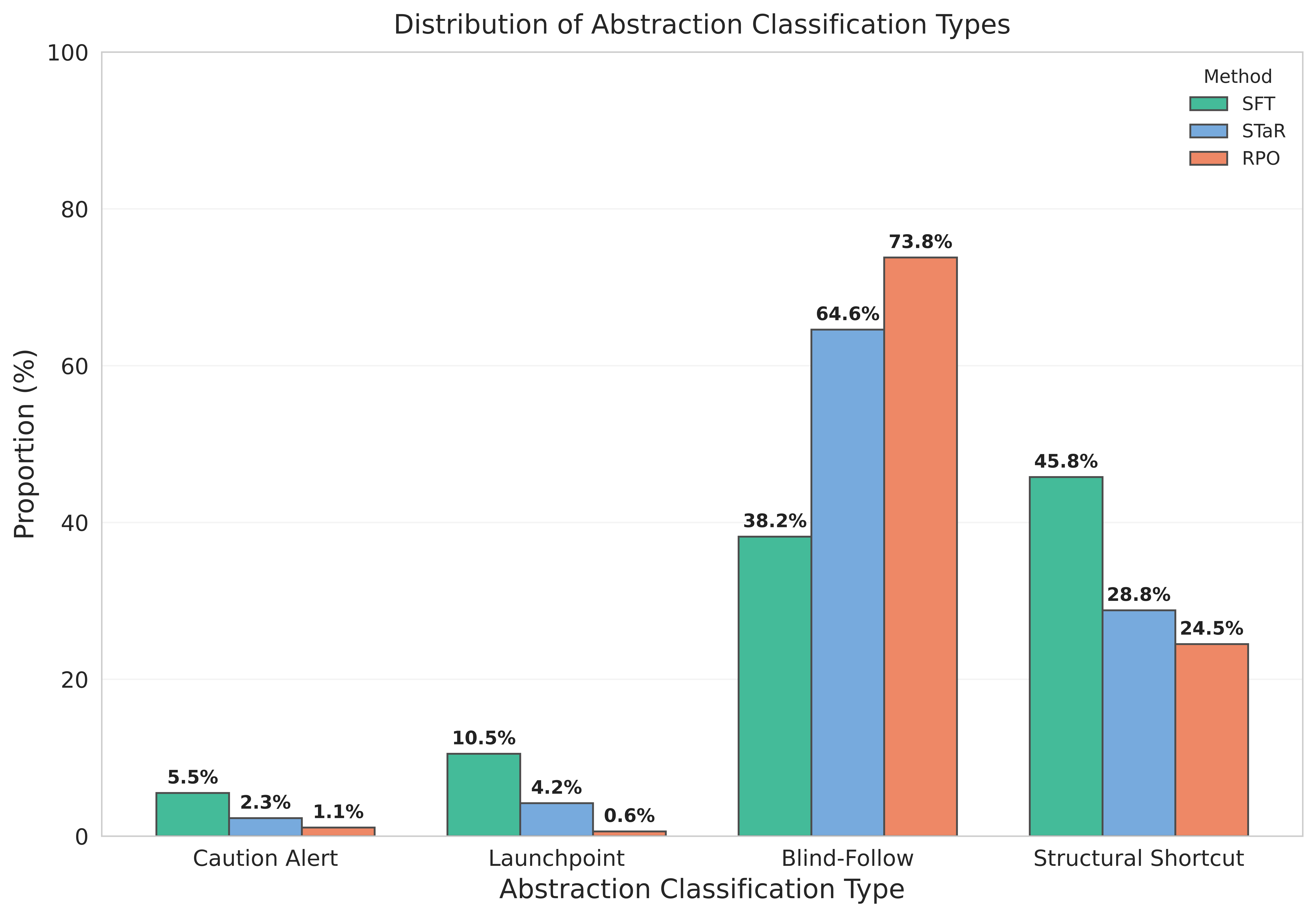}
  \vspace{-3mm}
    \caption{\footnotesize\textbf{Abstraction Categorization} \methodname{} produces a diverse characterization of abstractions, which we characterize by prompting \texttt{o4-mini}.}
  \vspace{-4mm}
  \label{fig:abstraction-classification}
\end{wrapfigure}
\textcolor{lightblue}{\emph{\textbf{Interpreting discovered abstractions.}}} As discussed in Appendix~\ref{app:abstractions_examples}, 
we classify each model‐generated abstraction into four categories for ease of interpretability: (1) \textbf{caution alert} that warns the solution generator to avoid a specific approach; (2) \textbf{productive launchpoint} that suggests strategic framings or problem reformulations that open high-potential solution paths; (3) \textbf{blind-follow trajectory} that prescribes repeatable, step-by-step procedures executable without further insight; and (4) \textbf{structural shortcut} that leverages abstract insights or invariants to collapse multiple reasoning steps into a single leap. In Figure~\ref{fig:abstraction-classification}, we show that after training via \methodname{}, the distribution over these categories shifts, with a notable increase in blind-follow abstractions, which a stronger reasoning model classifies as an effective reasoning path to a successful solution as seen in Appendix~\ref{app:abstractions_examples}.

\subsection{Prompt for Abstraction Classification}
We prompt \texttt{GPT-4o-mini} with the following prompt template to classify each abstraction into one of four categories.

\begin{promptbox}{Post-hoc abstraction classifier prompt}
\small
You are a abstraction classifier. You will be given a problem-solving heuristic or abstraction used for mathematical reasoning. Your task is to classify it into exactly one of the following mutually exclusive categories, based on the primary cognitive function the heuristic serves.

\medskip

(A) Caution alert: any abstraction that warns the reader to double-check a specific aspect of their solution or to not take a specific approach to the problem.

(B) Productive launchpoint: an early move or framing that opens up high-potential trajectories. Examples include clever reformulations or symmetries.

(C) Blind-follow trajectory: a description of a repeatable, sequential path that can be reliably followed to solve the problem. Examples include plug-and-play formulas that can be followed blindly, without insight. Do not choose this if further reasoning is required to solve the problem.

(D) Structural shortcut: a conceptual move that collapses multiple graph paths into a single jump via insight or abstraction. This can include introducing invariants.

(E) Other: a abstraction that does not fit into the above categories.

Give a 1-2 sentence explanation for your classification, and end your answer with exactly one of: (A), (B), (C), (D), or (E).

\medskip

---

abstraction: 

\{abstraction\}
\end{promptbox}

\subsection{Example for Short Abstraction}
\label{sec:short}

\begin{promptbox}{Examples of Short Abstractions}

\textbf{Question:} The banker's gain of a certain sum due 3 years hence at 
10\% per annum is Rs.~36. What is the present worth?

\textbf{Hint:} \\
- Consider the proportional relationship between the true discount and the present worth, and how the banker's gain relates to these terms. \\
- Think of the banker's gain as a component of the true discount and how it relates to the present worth through the interest rate and time period.

\vspace{1em}

\textbf{Question:} A train 125 m long passes a man running at 
15 km/hr in the same direction in which the train is going, in 10 seconds. 
What is the speed of the train?

\textbf{Hint:} \\
- Think about the speed difference between the train and the man, and how that difference relates to the time it takes for the entire train to pass by. \\
- Consider the difference in speeds as the key factor in determining the train's speed relative to the man, and use the time taken to pass the man to find this difference.

\end{promptbox}

\subsection{Example for Each Abstraction Category}
\label{sec:abs_category}
Below, we show examples of abstractions classified into the four categories above.

\begin{promptbox}{Examples of (A) Caution alert}
\small
<description>Always record forbidden values from denominators before and after manipulation. After solving the polynomial, discard any roots that make a denominator zero or that do not satisfy the original equation, to avoid extraneous solutions.</description>

<example>In the equation (x+2)/(2x–1) = x–3, 2x–1 cannot be zero (so x is not ½). If solving yields x=½ or any root that makes any denominator zero, reject it. Then verify the accepted roots in the original equation.</example>

\medskip

<description>Keep units consistent when moving between area and length or summing lengths. After extracting a length from an area (via square root), ensure subsequent arithmetic stays in the same unit to avoid scaling errors.
</description>

<example>If a square’s area is 10000 cm², its side is sqrt(10000) = 100 cm. To express in meters, convert 100 cm to 1 m. All later distances computed with that side length must be in meters to remain consistent.</example>
\end{promptbox}

\begin{promptbox}{Examples of (B) Productive launchpoint}
\small
<description>Translate comparative statements into algebraic equations using the chosen variables. Phrases like “twice as many” or “one less than” correspond to multiplication or addition/subtraction expressions. This step captures the core relationship in a solvable form.</description>

<example>If the problem states “Group A has twice as many as Group B,” write the equation x = 2y. For “Group B has three fewer than Group C,” you would write y = z - 3.</example>

\medskip

<description>Select one variable as a parameter (often setting it to 1 or keeping it symbolic) to express all other
variables in terms of it. This reduces the number of independent symbols and streamlines substitutions.</description>

<example>Given p/q = 3 and r/q = 2, choose q as the base variable. Write p = 3q and r = 2q, so all expressions involving p and r can be handled through q alone.</example>
\end{promptbox}
\vspace{0.5cm}
\begin{promptbox}{Examples of (C) Blind-follow trajectory}
\small
<description>Logarithms offer a streamlined way to compute floor-based digit counts: for y>0, the number of integer digits is floor(log10 y) + 1. Use this to handle arbitrary exponents without juggling large powers explicitly.</description>

<example>To count digits of y = $x^7$, compute d = floor(7 * log10 x) + 1. If x=2.5, then d = floor(7 * log10(2.5))+1 = 2+1 = 3 digits.</example>

\medskip

<description>The mean of a set equals its total sum divided by its number of elements. Use this to move between sums and averages when counts or totals are known. It works because “average” is defined as that ratio.</description>

<example>Suppose a subset has k items with mean m. Then its total sum is S = k·m. Conversely, if you know the sum S and the count k, the mean is m = S/k. For instance, if 5 items average to 10, their total is 5×10 = 50, and if you later learn the total is 60 for 6 items, the new mean becomes 60/6 = 10.</example>
\end{promptbox}

\vspace{0.5cm}

\begin{promptbox}{Examples of (D) Structural shortcut}
\small
<description>When the same distance appears in multiple geometric roles (e.g., as radius to a vertex and to a tangen
t point), express it in different algebraic forms and equate them. Solving the resulting equation produces the unknown variable, which then gives the desired length.</description>

<example>If r is both the distance from O to a vertex (r = sqrt[x² + (L/2)²]) and the distance from O to the tangent point (r = f(x)), set sqrt[x² + (L/2)²] = f(x). Solving this equation for x and back-substituting determines r explicitly, closing the geometric problem with an algebraic solution.</example>

\medskip

<description>Use the perimeter constraint a+b+c=P to eliminate one variable, e.g. set c=P-a-b, reducing the problem to two degrees of freedom. This simplification turns the three-variable Heron expression into a function of a and b alone, facilitating analysis or enumeration.</description>

<example>For a target perimeter P=10, one writes c=10-a-b. Substituting into Heron’s formula yields A(a,b)=sqrt[5 * (5-a) * (5-b) * (a+b-5)], which is now a two-variable function to study instead of three.</example>
\end{promptbox}

\begin{table}[htbp]
  \centering
  \begin{tabular}{lrrr}
    \toprule
    \textbf{Dataset} & \textbf{Zero-shot} & \textbf{Best} & \textbf{Average}\\
    & & \textbf{Abstraction} & \textbf{Abstraction}\\
    \midrule
    \small
    UCI Dry Bean & 0.00 & 0.65 & 0.51 \\
    Wikipedia Proteinogenic Acid & 0.22 & 0.78 & 0.58 \\
    UCI Student Performance & 0.25 & 0.45 & 0.28 \\
    UCI Website Phishing & 0.25 & 0.25 & 0.22 \\
    UCI Teaching Assistant Evaluation & 0.25 & 0.45 & 0.33 \\
    UCI Contraceptive Method Choice & 0.30 & 0.60 & 0.43 \\
    UCI Vertebral Column & 0.30 & 0.75 & 0.64 \\
    UCI Shill Bidding & 0.30 & 1.00 & 0.95 \\
    Kaggle Job Change & 0.30 & 0.85 & 0.83 \\
    UCI Caesarian Section & 0.38 & 0.75 & 0.64 \\
    Wikipedia Coin Face Value & 0.40 & 1.00 & 0.88 \\
    UCI Wine & 0.40 & 0.95 & 0.85 \\
    UCI Tic-Tac-Toe Endgame & 0.40 & 0.80 & 0.42 \\
    Kaggle Campus Placement & 0.40 & 0.85 & 0.72 \\
    Wikipedia Driving Championship Points & 0.40 & 1.00 & 0.74 \\
    UCI Mammographic Mass & 0.45 & 0.90 & 0.82 \\
    UCI Banknote Authentication & 0.45 & 1.00 & 0.78 \\
    Kaggle Engineering Placement & 0.50 & 0.85 & 0.79 \\
    RAFT One Stop English & 0.50 & 0.40 & 0.36 \\
    LegalBench Function of Decision Section & 0.54 & 0.72 & 0.61 \\
    Kaggle Entrepreneur Competency & 0.55 & 0.65 & 0.58 \\
    UCI Indian Liver Patient & 0.55 & 0.80 & 0.68 \\
    LegalBench International Citizenship Questions & 0.56 & 0.74 & 0.63 \\
    LegalBench Abercrombie & 0.56 & 0.80 & 0.67 \\
    Wikipedia Color Luminance & 0.60 & 1.00 & 1.00 \\
    RAFT Twitter Hate Speech & 0.60 & 0.90 & 0.76 \\
    Wikipedia Award Nomination Result & 0.64 & 1.00 & 0.76 \\
    UCI Car Evaluation & 0.65 & 0.75 & 0.64 \\
    Kaggle Water Potability & 0.65 & 0.50 & 0.38 \\
    Kaggle Travel Insurance & 0.65 & 0.70 & 0.59 \\
    UCI Internet Firewall & 0.70 & 1.00 & 0.97 \\
    RAFT ADE Corpus & 0.70 & 1.00 & 0.89 \\
    UCI Somerville Happiness Survey & 0.70 & 0.80 & 0.68 \\
    UCI Mushroom & 0.75 & 1.00 & 0.95 \\
    UCI Occupancy Detection & 0.80 & 1.00 & 0.92 \\
    Kaggle Stroke Prediction & 0.85 & 0.90 & 0.90 \\
    LegalBench Corporate Lobbying & 0.88 & 0.94 & 0.88 \\
    \midrule
    Average & 0.50 & 0.80 & 0.68 \\
    \bottomrule
  \end{tabular}
  \vspace{5pt}
  \caption{\label{tab:diverse_domains} Evaluation of abstractions on diverse collection of 37 domains. We sampled 10 abstractions by prompting \texttt{o4-mini}, and measure test set accuracy while prompting \texttt{GPT-4o-mini} with each abstraction. We report both the average performance of the 10 abstractions and the best abstraction. \textbf{We find that the average and best abstractions outperform standard prompting by 18.0\% and 30.0\% on average, respectively.}}
\end{table}

\end{document}